\documentclass{article}

\usepackage[final]{neurips_2021}

\usepackage[utf8]{inputenc} % allow utf-8 input
\usepackage[T1]{fontenc}    % use 8-bit T1 fonts
\usepackage{hyperref}       % hyperlinks
\usepackage{url}            % simple URL typesetting
\usepackage{booktabs}       % professional-quality tables
\usepackage{amsfonts}       % blackboard math symbols
\usepackage{nicefrac}       % compact symbols for 1/2, etc.
\usepackage{microtype}      % microtypography
\usepackage{xcolor}         % colors

\usepackage{graphicx}
\usepackage[ruled, lined, linesnumbered, commentsnumbered, longend]{algorithm2e}
\usepackage{subcaption}
\usepackage{natbib}
\setcitestyle{square,comma,numbers}
\usepackage{makecell}
\usepackage{diagbox}
\usepackage{amsmath}
\usepackage{wrapfig,lipsum,booktabs}
\usepackage{etoc}
\usepackage{multirow}
\usepackage{placeins}

\makeatletter
\newcommand{\printfnsymbol}[1]{%
  \textsuperscript{\@fnsymbol{#1}}%
}
\makeatother

\title{Learning to Delegate for Large-scale Vehicle Routing}

\author{%
    Sirui Li\thanks{Equal Contribution}\\
    MIT\\
    \texttt{siruil@mit.edu} \\
    \And
    Zhongxia Yan\printfnsymbol{1} \\
    MIT\\
    \texttt{zxyan@mit.edu} \\
    \AND
    Cathy Wu\\
    MIT\\
    \texttt{cathywu@mit.edu} \\
}

\begin{document}

\maketitle
\begin{abstract}
  Vehicle routing problems (VRPs) form a class of combinatorial problems with wide practical applications. While previous heuristic or learning-based works achieve decent solutions on small problem instances of up to 100 cities, their performance deteriorates in large problems. This article presents a novel learning-augmented local search framework to solve large-scale VRP. The method iteratively improves the solution by identifying appropriate subproblems and \textit{delegating} their improvement to a black box subsolver. At each step, we leverage spatial locality to consider only a linear number of subproblems, rather than exponential. We frame subproblem selection as regression and train a Transformer on a generated training set of problem instances. Our method accelerates state-of-the-art VRP solvers by 10x to 100x while achieving competitive solution qualities for VRPs with sizes ranging from 500 to 3000. Learned subproblem selection offers a 1.5x to 2x speedup over heuristic or random selection. Our results generalize to a variety of VRP distributions, variants, and solvers.
\end{abstract}

\section{Introduction}\label{sec:intro}
Vehicle routing problems (VRPs) have enjoyed ample applications in logistics and ride-hailing services~\citep{laporte2009fifty} around the world for decades. While determining the optimal solution to a VRP is NP-hard~\citep{lenstra1981complexity}, there have been numerous attempts to solve VRPs both exactly and approximately: provable algorithms have been designed for specific problem instances up to size 130~\citep{lysgaard2004}, and powerful heuristic solvers such as LKH-3~\citep{helsgaun2017} and HGS~\cite{vidal2012hybrid, vidal2020hybrid} find good solutions in practice for problems of size more than 1000. However, heuristics methods often suffer from inflexibility due to extensive hand-crafting and the heavy computational burden from lengthy iterative procedures, as in the case of LKH-3. For example, LKH-3 takes more than an hour to solve a size 2000 CVRP instance (Table~\ref{tab:models}), which is impractical for applications such as large-scale courier or municipal services.

More recently, machine learning methods inspired by the Pointer Network~\citep{vinyals2015pointer} provide alternatives to traditional solvers: the learning-based methods greatly reduce computation time while maintaining decent solution quality on small problem instances (less than 100 cities), by training on diverse sets of problem distributions either via supervised~\citep{vinyals2015pointer} or reinforcement learning~\citep{nazari2018reinforcement, kool2018attention}. However, these methods remain difficult to scale, and few report results on problems of size more than 200.

Our work aims to address scalability by learning to identify smaller subproblems which can be readily solved by existing methods. Our learned subproblem selector guides the problem-solving process to focus local improvement on promising subregions. While there exists a combinatorial number of subproblems that can be selected at each step, we leverage spatial locality commonly found in combinatorial optimization problems to restrict the selection space size to be linear in the problem size. Intuitively, objects far away from each other generally have very small influence on each other's solution and are likely to be in different routes, so they should not be part of the same subproblem. The greatly reduced search space enables us to feasibly train an attention-based subproblem selector.

Our framework combines the advantages of learning and heuristics: our network identifies promising subproblems to improve upon, dramatically speeding up solution times. Using a competitive subsolver on subproblems, we achieve good solution quality without the high computational costs of running the subsolver on large problem instances. In summary, our contributions are:
\begin{itemize}
    \item We propose \textit{learning-to-delegate}, a learning-based framework for solving large-scale VRPs by iteratively identifying and solving smaller subproblems.
    \item Despite the high dimensionality and NP-hardness of subproblems, we design a Transformer architecture that effectively predicts the subsolver's solution quality for a subproblem.
    \item With extensive validation, we show that learning-to-delegate offers significant speedups and/or objective improvements over both its base subsolver and random (or heuristic) subproblem selection, for a variety of VRP variants, distributions, and solvers.
\end{itemize}

\section{Preliminary: Capacitated Vehicle Routing Problems (CVRP)}
In CVRP, there is a depot node $0$ and city nodes $\{1, ..., N\}$. Each city node $i$ has demand $d_i$ to fulfill. A vehicle with capacity $C$ starts and ends at the depot and visits a route of city nodes such that the sum of city demands along the route does not exceed $C$, after which the vehicle starts a new route again. The objective is to find a valid solution minimizing the solution cost. We define the following:
\begin{itemize}
    \item Route: a sequence of nodes, where the first and last node are the depot $0$, and the rest are city nodes. In a valid route, the sum of demands of the city nodes does not exceed $C$.
    \item Route cost: the sum of edge costs for the sequence of nodes. For an edge from node $i$ to node $j$, the edge cost is the Euclidean distance between node $i$ and $j$.
    \item Solution: a \textit{feasible} solution consists of a set of valid routes visiting each city exactly once.
    \item Solution cost: the sum of route costs for all routes in the solution. An \textit{optimal} solution is a feasible solution with the minimum solution cost.
    \item Subproblem: a CVRP consisting of the depot, a subset of cities, and corresponding demands.
\end{itemize}
\section{Related Work}\label{sec:related_work}
\paragraph{Classical methods.}
Heuristics for solving combinatorial optimization problems have been studied for decades. The most powerful methods, such as local search~\citep{aarts2003local}, genetic algorithms~\citep{sivanandam2008genetic}, and ant colony methods~\citep{dorigo2006ant}, involve iteratively improving the solution in a hand-designed neighborhood. For example, move, swap~\citep{wu2016optimizing}, and 2-opt~\citep{croes1958method} are well-known heuristics for the traveling salesman and vehicle routing problems. The competitive VRP solver LKH-3 \citep{helsgaun2017} uses the Lin–Kernighan heuristic \citep{lin1973effective} as a backbone, which involves swapping pairs of sub-routes to create new routes, whereas the CVRP solver HGS~\cite{vidal2012hybrid, vidal2020hybrid} uses a hybrid genetic and local search procedure to achieve state-of-the-art solution qualities on problems up to size 1000. While LKH-3\footnote{We refer to the LKH-3 code at \url{http://webhotel4.ruc.dk/~keld/research/LKH-3/}.} tackles a variety of VRP variants, the publicly available implementation of HGS~\footnote{We refer to the HGS code at \url{https://github.com/vidalt/HGS-CVRP}.} only solves CVRP.

For large problems, low-level heuristics are combined with \textit{meta-heuristics}, including Tabu Search with Adaptive Memory~\citep{taillard2001adaptive}, guided local search~\citep{voudouris2003guided}, and Large Neighborhood Search~\citep{shaw1998using}. The inspiration for our work derives from Partial OPtimization Metaheuristic Under Special Intensification Conditions (POPMUSIC), which iteratively optimizes problem subparts and has been used to solve problems as diverse as map labeling \citep{laurent2009point}, berth allocation \citep{lalla2016popmusic}, and $p$-median clustering \citep{taillard2003heuristic}.

Despite the promise of the POPMUSIC framework for large-scale combinatorial optimization, the impact of certain design choices is not well understood, such as the subproblem selection ordering and size~\citep{Taillard2018}. For example, it is not clear how to meaningfully order the subproblems. One early work in this direction demonstrated that a last-in-first-out stack order performs better than random~\citep{alvim2013popmusic}. Our work provides a natural approach to order the subproblems based on predicted improvement.

\paragraph{Deep learning methods.}
Recently, there has been a surge of interest in training deep neural networks (DNN) to solve combinatorial problems. As observed by~\citet{NEURIPS2020_f231f210}, most methods fall into one of the following two categories: 1) \textit{construction methods}~\citep{nazari2018reinforcement, kool2018attention}, where an autoregressive model such as the Pointer Network~\citep{vinyals2015pointer} directly outputs a solution, and 2) \textit{improvement methods}~\citep{NEURIPS2019_131f383b, hottung2019neural}, where the DNN iteratively performs local updates to the solution, resembling local search.

These methods are approaching the solution quality of LKH-3~\citep{helsgaun2017} on small problem instances. For example, \citet{NEURIPS2020_f231f210} extends a construction approach~\citep{kool2018attention} to encourage more diverse output samples on $N \leq 100$ VRPs. Among improvement methods, \citet{Lu2020A} learn a meta-controller to select among a set of local search heuristics, marking the first learning-based approach to outperform LKH-3 on $N \leq 100$ VRPs. Despite these successes, learning-based approaches for large-scale VRPs are poorly understood.
In addition, all of the aforementioned methods are trained using deep reinforcement learning; for large problems, trajectory collection becomes prohibitively expensive.

\paragraph{Scaling up learning methods.}
A few recent works have begun to investigate scaling of learned networks for NP-hard graph problems. For example, \citet{ahn2020learning} propose an iterative scheme called \textit{learning-what-to-defer} for maximum independent set. At each stage, for each node in the graph, the network either outputs the solution for the node or defers the determination to later stages. \citet{song2020general} proposes an imitation-learning-based pipeline for Integer Linear Programs (ILP), where at each stage they partition all variables into disjoint subsets, and use the Gurobi \citep{gurobi} ILP solver to solve the partitions. Due to differences in graph structure, our work presents a more natural scheme to handle VRP constraints and structures. %an expanded discussion of other related learning-based approaches to this problem, 
A few works attempt to incorporate learning to decompose large-scale VRPs~\cite{bosman2006computationally, ventresca2013predicting, poullet2020leveraging}. However, the decomposition approaches proposed appear to be experimentally less effective than ours.

\section{An Iterative Framework for VRPs}
\begin{figure*}
  \centering
  \hbox{\hspace{5em} \includegraphics[width=0.9\linewidth,page=3]{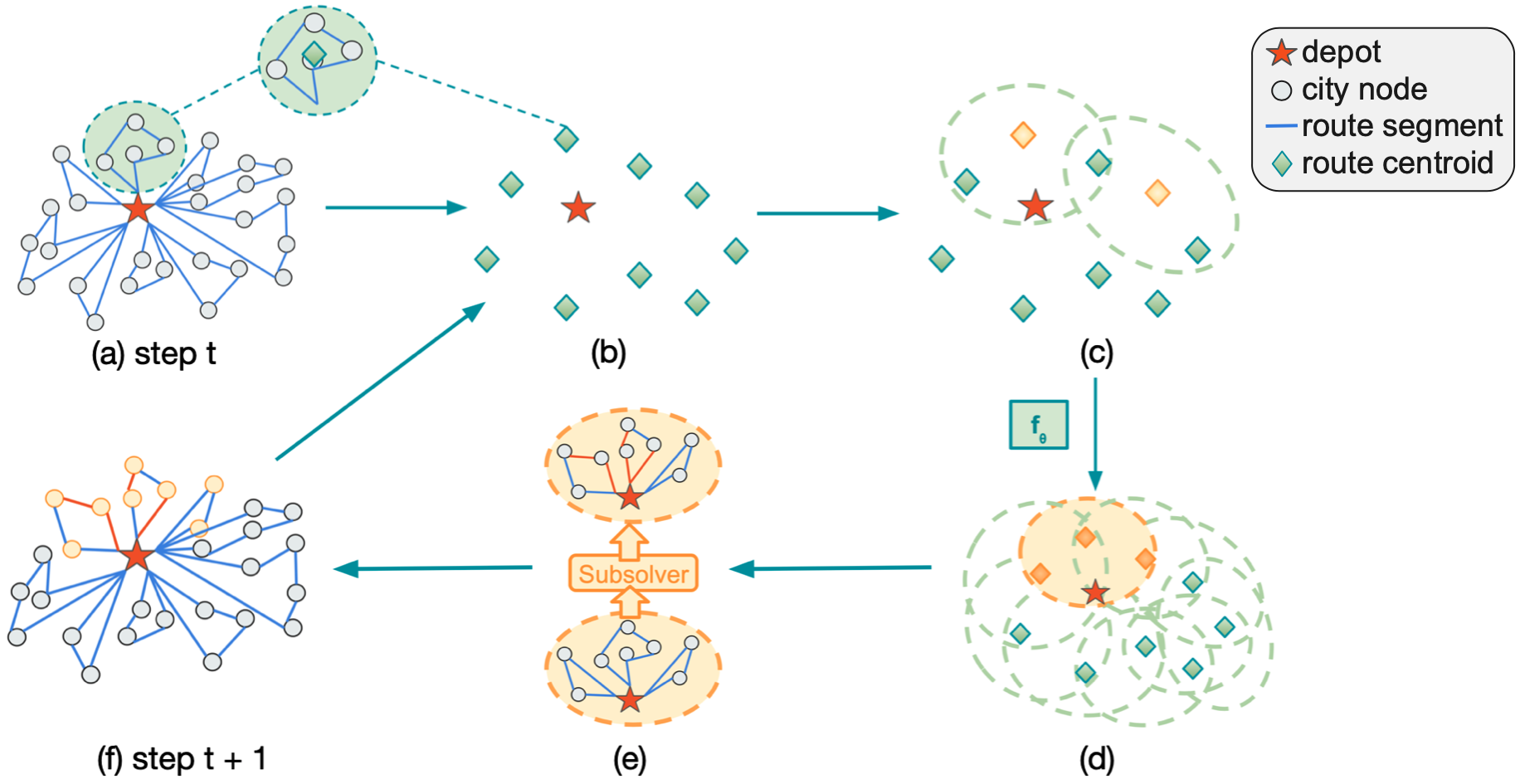}}
  \caption{\textbf{Our iterative framework for VRPs}. (a) At each time step, we start with a current solution $X$. Circles are city nodes, blue lines are route segments, and the red star is the depot.
(b) We aggregate each route by taking the centroid of all city nodes in the route.
(c) For each route, we define a corresponding subproblem as the $k$-nearest neighbors of the route. Two such subproblems with $k=3$ are shown. 
(d) Our subproblem selector selects a subproblem $S$ (yellow).
(e) We feed $S$ into the subsolver to get a new subsolution $X_S'$. The red edges are updated by the subsolver.
(f) We update $X$ to new solution $X'$ with $X_S'$, then repeat (b)-(f).}\label{fig:pipeline}
\end{figure*}

While CVRPs are NP-hard and thus require worst-case exponential time in the problem size to solve optimally, typical CVRPs exhibit structure that practical solvers may exploit. We hypothesize that in such situations, the larger problem can be efficiently approximately solved as a sequence of smaller subproblems, which can be \textit{delegated} to efficient subsolvers. To test this hypothesis, we propose a learning-based, iterative framework with two components: 1) a \textit{subsolver} capable of solving a small problem instance (exactly or approximately), and 2) a \textit{learned model} for identifying suitable subproblems within a larger problem to delegate to the subsolver.

\begin{algorithm}
\label{alg:main}
\SetAlgoLined
\KwIn{Problem instance $P$, initialized solution $X$, subproblem selector $f_\theta$, subsolver $Subsolver$, number of steps $T$, parameter $k$ denoting the size of subproblems}
 \For{Step t = 1: T}{
    $\mathcal S_{k,\textrm{local}} \leftarrow \text{ConstructSubproblems}(P, X, k)$\\
    $S \leftarrow f_\theta(\mathcal S_{k,\textrm{local}})$\\
    $X_S' \leftarrow \text{Subsolver}(S)$\\
    $X \leftarrow X_S' \cup X_{P \setminus S}$\\
 }
\caption{Learning to Delegate}
\end{algorithm}

We illustrate our iterative framework in Figure~\ref{fig:pipeline}, which takes in a problem instance $P$ with a feasible initial solution $X_0$. At each step, given the current solution $X$, we select a smaller subproblem $S \subset P$ with our learned model $f_\theta$ then apply the subsolver to solve $S$; we then update $X$ with the new solution for $S$. To maintain feasibility after an update, we restrict $S$ to be the set $\mathcal S$ of visited cities of a \textit{subset of routes} from $X$. Since routes with cities in $P \setminus S$ remain valid routes, we obtain a new feasible solution $X' = X_S' \cup X_{P\setminus S}$, where $X_S'$ is the subsolution for $S$ and $X_{P \setminus S}$ consists of unselected routes from $X$. Intuitively, a strong $f_\theta$ should identify a subproblem $S$ such that the subsolver solution $X_S'$ results in a large improvement in objective from $X$ to $X'$.

\subsection{The Restricted Subproblem Selection Space}
\label{sec:subproblem_space}

As the number of routes in $P$ is $R = O(N)$, the cardinality of the selection space $\mathcal S$ is $O(2^R)$, which is exponential in the problem size $N$ and difficult for learned models to consider. If we restrict each subproblem to cities from exactly $k$ routes, there are still ${R \choose k} = O(R^k)$ subproblems to consider. Therefore we further restrict selection to subproblems with spatial locality. As shown in Figure~\ref{fig:pipeline}(d), we only consider subproblems from $\mathcal S_{k,\textrm{local}}$, where each subproblem is centered around a particular route $r$ and contains the $k$ routes whose centroids have the smallest Euclidean distance to the centroid of $r$. In this way, we reduce the selection space to $|\mathcal S_{k,\textrm{local}}| = R = O(N)$ from $|\mathcal{S}| = O(2^N)$. In Algorithm 1, we refer to this restriction as $\text{ConstructSubproblems}$. Our restriction to a local selection space is motivated by the fact that many combinatorial optimizations problems have inherent spatial locality, i.e. problem entities are more strongly affected by nearby entities than faraway entities. Earlier heuristical methods such as POPMUSIC~\citep{taillard2002popmusic} leverage similar spatial locality.

\section{Learning to Delegate}
\label{sec:learningtodelegate}
In this section, we discuss criteria for selecting subproblems, and how to train the subproblem selector.  

\paragraph{Improvement as the criteria for subproblem selection.}
Given a selected subproblem $S$ with a current solution $X_S$ on the subproblem, we obtain from LKH-3 a new solution $X'_S$ on the same subproblem (step $(d)$ to $(e)$ in Figure \ref{fig:pipeline}). We then define the \textit{immediate improvement} \begin{equation}
\delta(S) = c(X_S) - c(X'_S)
\label{eq:immediate_improvement}
\end{equation}
where $c(X_S)$ is the total cost of subsolution $X_S$ and $\delta(S)$ is the improvement in the solution cost. In this way, the sum of improvements along $T$ steps is the total improvement in solution quality. As we empirically find that providing the previous subsolution $X_S$ to the subsolver may trap the subsolver in a local optimum, we withhold $X_S$ from the solver and thus may see non-positive improvement $\delta(S) \leq 0$, especially after many steps of subproblem selection. At test time, to avoid worsening the objective, we adopt a hill-climbing procedure such that when $\delta(S) \leq 0$, we keep $X_S$ instead of $X_S'$ (Figure~\ref{fig:pipeline}, step $(e)$). With proper masking, we avoid selecting the same non-improving subproblem $S$ again. The hill-climbing and masking procedures are applied to both our subproblem-selection network $f_\theta$ and the three heuristic selection rules, described in \ref{sec:setup}, to maintain fair comparisons. 

\paragraph{Subproblem selection.} The goal of our subproblem selector is to select the subproblem leading to the best \textit{immediate} improvement. While our subproblem selection does not directly optimize for the total improvement, we observe that (1) our subsolver may perform many low-level operations internally, so high-level problem selection may still benefit greatly from maximizing the immediate improvement, and (2) numerous reinforcement learning approaches to VRP~\citep{NEURIPS2019_12e59a33,dai2017learning} choose a small discount factor such as $\gamma=0.25$ when optimizing a multi-step objective $\sum\limits_{t=1}^{T}\gamma^t \delta_t$, as doing so encourage faster convergence. Moreover, our subproblem selector may instead be trained on multi-step search data to select subproblems offering the best multi-step improvement.

\paragraph{Ground-truth labels.} Our restricted selection space allows us to enumerate all possible subproblems at each step. In a typical large scale CVRP instance with $N = 2000$ cities, a solution consists of roughly $R = 200$ routes, so the size of the selection space is 200. We obtain the immediate improvement $\delta(S)$ by running the subsolver on each subproblem $S \in \mathcal S_{k,\textrm{local}}$. Although our enumeration strategy is feasible for generating training data, it is much too slow to execute on a test CVRP instance. However, if our subproblem selector can predict the best immediate improvement at test time (that is, without running the subsolver on multiple subproblems), then we can combine the best of both worlds to obtain a fast and accurate selection strategy.

\begin{figure}
  \centering
  \includegraphics[width=0.9\linewidth,page=3]{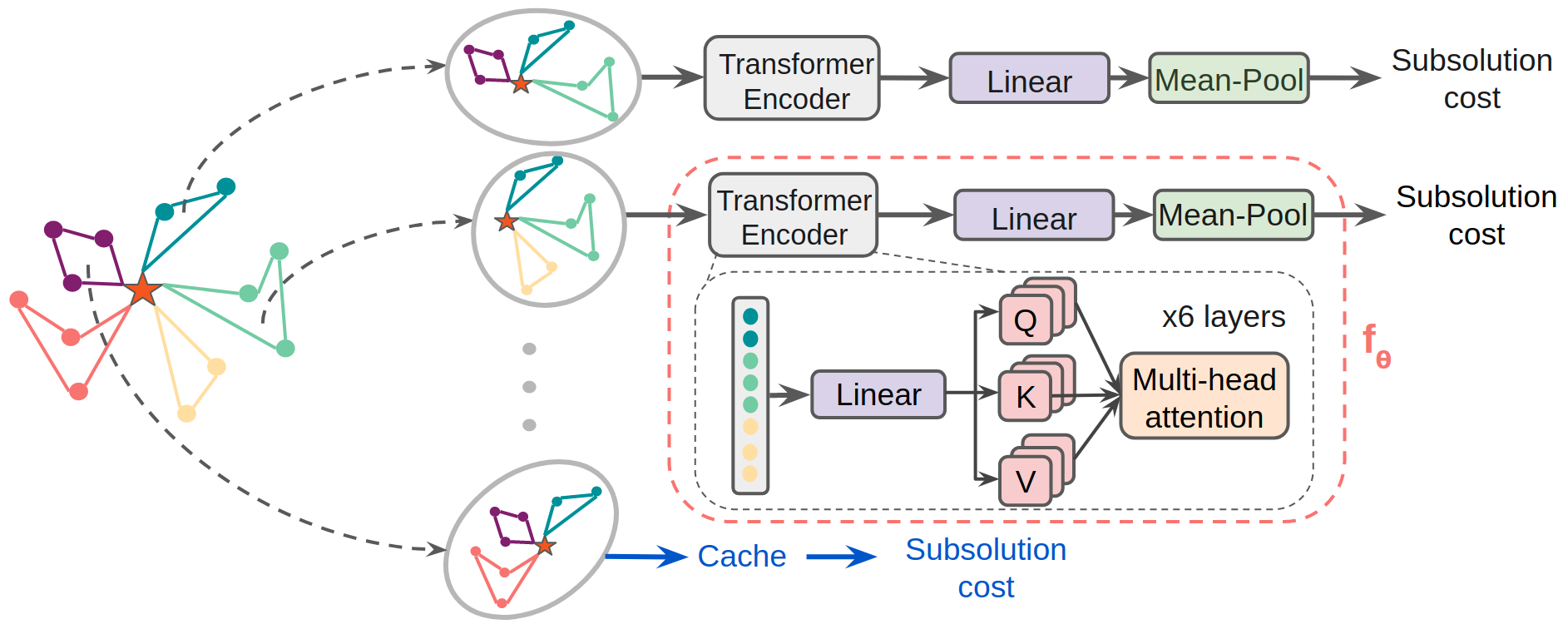}
  \caption{\textbf{Our Transformer architecture}. At each step of our framework (Figure~\ref{fig:pipeline}(c)), we featurize subproblem $S \in \mathcal{S}_{k,\text{local}}$ into a column vector of unordered cities. We apply the Transformer encoder with multiple multi-head attention layers and a final linear layer, before mean-pooling over the cities to generate the predictions $f_\theta(S)$, which we fit to the subsolution cost $c(X_S')$. When possible, we retrieve a previously predicted subproblem from a cache to minimize computation.}\label{fig:architecture}
\end{figure}

\paragraph{Selection strategies: regression vs classification.} Given a labeled dataset, we may treat the task of identifying the best subproblem as either regression or classification. In the context of imitation learning with a greedy expert, the former learns a \textit{value function} while the latter learns a \textit{policy}. Due to space limitation, we focus discussion on regression and reserve comparison with classification for Appendix~\ref{appsec:classification}. The regression-based subproblem selector uses a trained $f_\theta$ to predict the subsolution cost $c(X_S')$; we then simply compute $\arg\max_S c(X_S) - f_\theta(S)$ to select the best subproblem.

\paragraph{Network architecture.} We define $f_\theta$ with a Transformer encoder architecture~\citep{vaswani2017attention}. The input representing each subproblem $S$ is the unordered set of featurized cities in $S$. The features for each city consist of the demand $d$ and the location of the city $(x, y)$ relative to the depot. As we do not feed the existing subsolution $X_S$ to the subsolver, the dense multi-head attention mechanism does not need to be modified to take the routes in $X_S$ into account. The output of the Transformer encoder is fed into a linear layer then mean-pooled to obtain the scalar prediction $f_\theta(S)$. In Appendix~\ref{appsec:architecture_ablation}, we perform an ablation study with simpler architectures.

\paragraph{Loss function.} We empirically find mean squared error to be less stable than Huber loss \citep{huber1992robust}, which we set as our loss function
\begin{equation}
    L(\theta; S) = \begin{cases}
    \frac 1 2 (f_\theta(S) - c(X_S'))^2, & \textrm{if } |f_\theta(S) - c(X_S')| \leq 1\\
    |f_\theta(S) - c(X_S')| - \frac 1 2, & \textrm{otherwise}
    \end{cases}
\end{equation}

\section{Experiments and Analysis}
\label{sec:experiment}
\begin{figure*}
  \centering
  \includegraphics[width=0.9\linewidth]{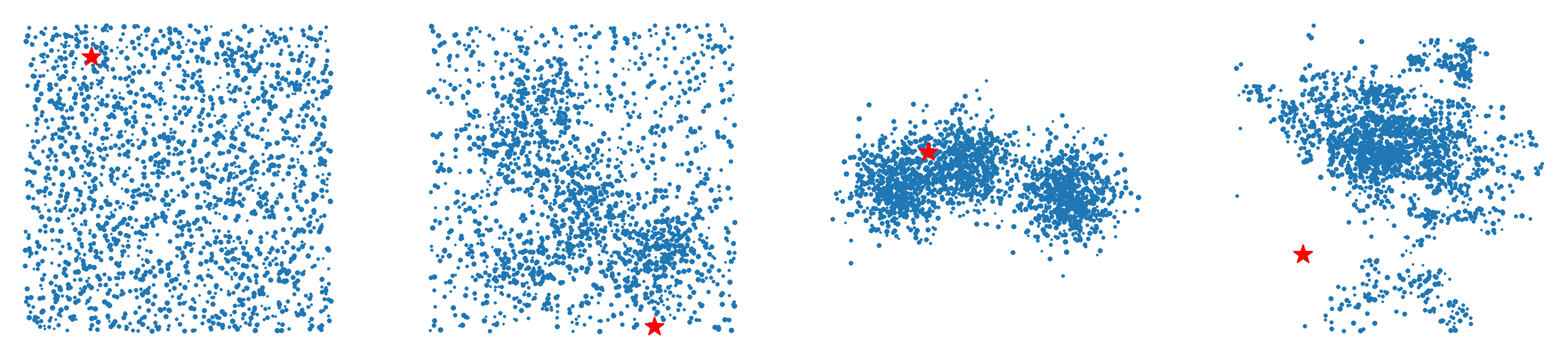}
  \caption{\textbf{Instances from $N = 2000$ CVRP distributions}. From left to right: instance from uniform, mixed ($n_c = 7$ cluster centers), clustered ($n_c = 3$ cluster centers), and real-world. The red star is the location of the depot, while blue dots are cities sized proportional to demand.}\label{fig:dists}
\end{figure*}

We illustrate the CVRP distributions considered in our work in Figure~\ref{fig:dists}. We perform extensive experiments to evaluate our learning framework, aiming to answer the following questions:
\begin{enumerate}
    \item \textbf{Uniform distribution}. How does our method compare with baselines, in terms of solution time and quality, on problems with uniformly distributed cities?
    \item \textbf{Clustered distributions}. How does our method perform on problems with clustered cities?
    \item \textbf{Out-of-distribution}. Can our model generalize, such as to larger or real-world instances?
    \item \textbf{VRP variants}. Can our method address more sophisticated VRPs? E.g.,  CVRP with Time Windows (CVRPTW)~\citep{solomon1987algorithms} or VRP with Mixed Pickup and Delivery (VRPMPD) ~\citep{salhi1999cluster}.
    \item \textbf{VRP solvers}. Can our method be adapted to leverage other VRP subsolvers?
\end{enumerate}

We reserve additional ablations on subproblem selection as classification, effect of subproblem size $k$ and discussion of asymptotic behavior, subproblem selection with the HGS subsolver, effect of weaker initialization methods, and comparison with simpler architectures for Appendix~\ref{appsec:ablation}.

\subsection{Setup}\label{sec:setup}
We briefly describe experimental setup in the main text and defer full details to Appendix~\ref{appsec:uniform_setup}. Given a particular distribution of VRP, we generate separate sets of training, validation, and test problem instances. Unless otherwise stated, our validation and test sets contain 40 instances each. For each problem instance, we generate a rough initial solution by partitioning it into disjoint subsets of cities and briefly running the subsolver on each subset. Due to its compatibility with many VRP variants, we use LKH-3~\citep{helsgaun2017} as the subsolver for all VRP distributions unless otherwise stated.

We generate the training set by running our iterative framework and selecting subproblems by enumerating the subsolver on all $S \in \mathcal S_{k, \text{local}}$. As many subproblems remain unchanged from $X$ to $X'$, we use previously cached subsolutions when possible instead of re-running the subsolver. While generation times differ depending on several factors, typically it takes less than 10 hours with 200 Intel Xeon Platinum 8260 CPUs to generate a training set of 2000 instances.

To avoid training multiple models for different problem sizes $N$, we train a single model for each VRP distribution with combined data from multiple $N \in \{500, 1000, 2000\}$. Training takes at most 8 hours on a single NVIDIA V100 GPU. To exploit symmetry in problem instances, we apply rotation and flip augmentation at training time. To evaluate trained selectors on a problem instance, we run the iterative selection framework on a single CPU, with disabled multithreading and no GPU.

We select the best hyperparameters via manual search based on validation set performance. We record final results on the test set as the mean and standard error over 5 runs with different random seeds.

\paragraph{Baselines.} 
%A key baseline from the literature is \textbf{LKH-3}~\citep{helsgaun2017}.
By default, as we use LKH-3 as the subsolver, we run LKH-3 on the full problem instances with our initialization for $30000$ local update steps, which takes 2-3 hours on average for $N = 2000$. If using HGS as the subsolver, we also run HGS on the full problem instances for the same amount of time as our LKH-3 baseline. These baselines allow us to compute the speedup of our framework over the subsolver alone. We also compare against OR Tools~\citep{ortools}, another open source heuristic VRP solver employing iterative search, terminating runs at 15 minutes, as OR Tools stops improving the solution within this time for all instances. 

We include results for previous learning methods AM~\citep{kool2018attention} and NeuRewriter~\citep{NEURIPS2019_131f383b}. These do not outperform LKH-3 even on small problem sizes, and learning methods have had more difficulty generalizing to larger instances. In fact, these methods are trained on problems of size $N \leq 100$, and we find that they yield poor solutions on $N \geq 500$ without architecture modifications and extensive re-training. The AM and NeuRewriter results demonstrate the difficulty of scaling up previous learning methods. We do not initialize OR-Tools, AM, and NeuRewriter because we empirically find that these methods have limited solution capability and do not improve our decent initialization.

To validate our subproblem selector's ability to identify promising subproblems, we design three additional baselines that employ our iterative framework using \textit{hand-crafted heuristics} to select subproblems. The three heuristics that we use are: (1) Random, selecting subproblem $S$ from $\mathcal S_{k, \text{local}}$ uniformly; (2) Count-based, which avoids repetitive selections by selecting the subproblem centered at the route whose city nodes have been selected cumulatively the least often in previous steps; and (3) Max Min Distance, which encourages coverage of the entire problem instance by selecting subproblems with the maximum distance from the nearest centroid of previously selected subproblems. We run the heuristic baselines with the same setup as our learned subproblem selector.

\paragraph{Metrics.} We refer to two metrics to compare our method against baseline methods.
\begin{enumerate}
    \item \textbf{Improvement over method X}: at a specified computation time, the improvement of method Y over method X is the total improvement of method Y minus that of method X.
    \item \textbf{Speedup over method X}: at a specified solution quality that method X attains, the speedup of method Y over method X is the computation time required for method X to attain the solution quality divided by the time for method Y to attain the solution quality.
\end{enumerate}

We define \textit{95\% solution quality} of running a method X over a computation time as the solution quality with 95\% of the total improvement. We report speedup at 95\% solution quality of a method X because X may take a disproportionate amount of time on the last 5\% of improvement; reporting speedup at 100\% solution quality inflates the speedup.

\begin{table}
    \caption{\textbf{Performance and computation time for uniform CVRP}. For problem instance sizes $N \in \{500,1000, 2000\}$, we report the objective values (\textit{lower} is better) of our method and baseline methods, averaged across all instances in the test set. Note that the cost is the total distance of routes in the solution. LKH-3 (30k) runs LKH-3 for 30k steps to near convergence, while LKH-3 (95\%) is the 95\% solution quality. Random, Count-based, Max Min Distance, and Ours (Short) run until matching LKH-3 (95\%) in solution quality, with the speedup reported in parentheses, while Ours (long) runs for twice amount time as Ours (Short).\vspace{0.1cm}}\label{tab:models}
    \centering
    \begin{tabular}{rcccccc}
      \toprule % from booktabs package
       \bfseries  & \multicolumn{2}{c}{\bfseries N = 500} &  \multicolumn{2}{c}{\bfseries N = 1000} &   \multicolumn{2}{c}{\bfseries N = 2000}\\
       \bfseries  Method & \bfseries Cost & \bfseries Time & \bfseries Cost & \bfseries Time  & \bfseries Cost & \bfseries Time \\
      \midrule % from booktabs package
      LKH-3 (95\%) & 62.00  &  4.4min & 120.02 & 18min & 234.89 & 52min \\
      LKH-3 (30k) & 61.87  &  30min & 119.88 & 77min & 234.65 & 149min \\
      \hline
      OR Tools & 65.59  &  15min & 126.52 & 15min & 244.65 & 15min \\
      AM sampling & 69.08 & 4.70s & 151.01 & 17.40s & 356.69 & 32.29s\\
      AM greedy & 68.58 & 25ms  & 142.84 & 56ms &  307.86 & 147ms\\
      NeuRewriter & 73.60 & 58s & 136.29 & 2.3min & 257.61 & 8.1min\\
      \hline
      Random & 61.99  &  71s (3.8x) & 120.02 & 3.2min (5.5x) & 234.88 & 6.4min (8.0x) \\
      Count-based  & 61.99  & 59s (4.5x) & 120.02 & 2.1min (8.2x) & 234.88 & 5.3min (10x) \\
      Max Min & 61.99  & 59s (4.5x) & 120.02 & 2.5min (7.0x) & 234.89 & 5.2min (10x)\\
      \hline
      Ours (Short) & 61.99  & 38s (7.0x)  & 119.87 & 1.5min (12x) & 234.89 & 3.4min (15x) \\  
      Ours (Long) & \textbf{61.70}  & 76s & \textbf{119.55} & 3.0min & \textbf{233.86} & 6.8min \\  \bottomrule % from booktabs package
    \end{tabular}
\end{table}
\subsection{Uniform CVRP Distribution}
\label{sec:uniform}
As seen in Table~\ref{tab:models}, our method achieves the best performance for all problem sizes, matching LKH-3's solution quality with more than 7x to 15x less computation time and offering even more improvements with longer computation time. Although we need to evaluate all $R = O(N)$ subproblems of the initial solution with our subproblem selector, subsequent per-step computation time of our method is mostly independent of the problem size $N$ since we only evaluate changed subproblems.

Running LKH-3 for 30k local update steps achieves superior performance to all previous other heuristic and learning-based baselines. Its solution quality scales well to large problem sizes, yet the solution time is significantly longer. Previous learning-based methods, though fast, result in much worse solution qualities. Our heuristic baselines Random, Count-based, and Max Min Distance demonstrate that our iterative framework, even without the learned subproblem selector, may achieve over 5x to 10x speedup over LKH-3. Nevertheless, our results demonstrate that learning the subproblem selector may offer an additional 1.5x speedup over non-learning heuristics.

\begin{figure*}
  \centering
  \includegraphics[width=\linewidth]{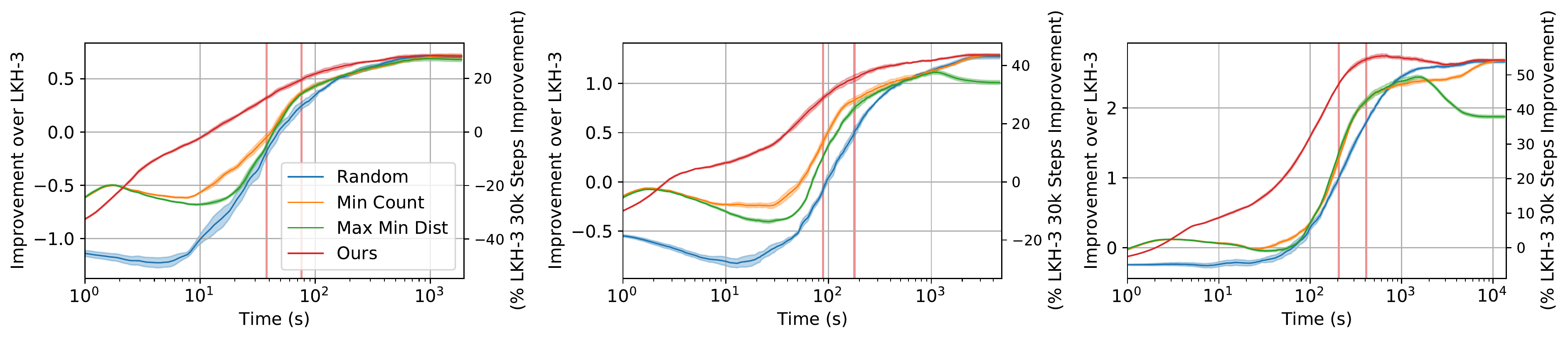}
  \caption{\textbf{Improvement over LKH-3 for uniform CVRP}. The x-axis is the computation time and extends until LKH-3 has completed 30k steps. The vertical lines represent the computation times of Ours (Short) and Ours (Long) from Table~\ref{tab:models}. The three subplots correspond to $N=500$ (left), $N=1000$ (middle), and $N=2000$ (right).}\label{fig:uniform_improvement}
\end{figure*}

\begin{figure*}
  \centering
  \includegraphics[width=\linewidth]{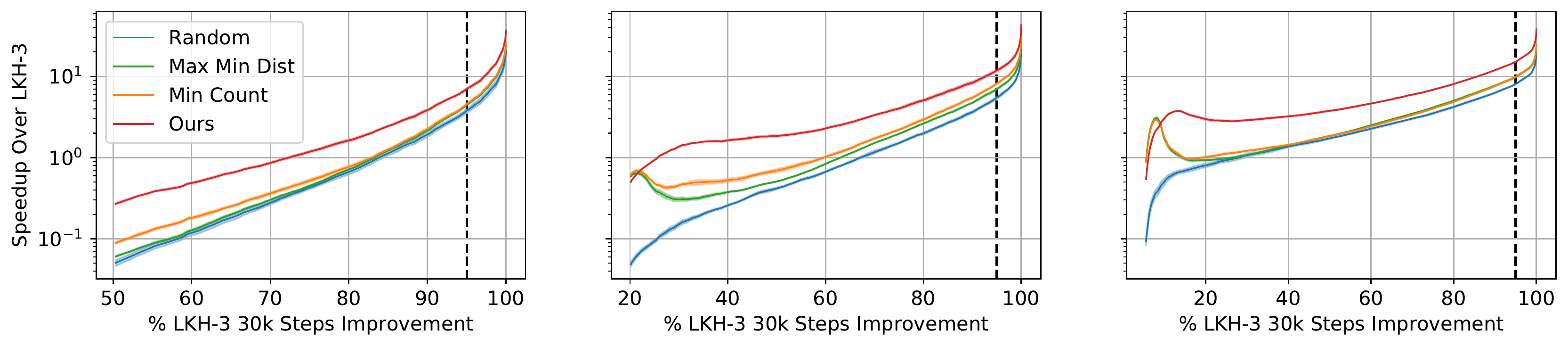}
  \caption{\textbf{Speedup over LKH-3 for uniform CVRP}. The x-axis is the solution quality attained, measured as a percentage of the LKH-3's maximum improvement. The dashed vertical line represents the 95\% solution quality used to compute the speedup, as mentioned in Table~\ref{tab:models}. The three subplots correspond to $N=500$ (left), $N=1000$ (middle), and $N=2000$ (right).}\label{fig:uniform_speedup}
\end{figure*}
In Figure~\ref{fig:uniform_improvement}, we demonstrate the solution quality of our method and baselines compared to LKH-3. We see that, compared to baseline methods, the learned subproblem selector obtains the best solution quality when run for a reasonable amount of time. The improvement of most methods based on our iterative framework converge when run for an excessive amount of time; this is unsurprising, as Random and other baselines are eventually able to select all subproblems offering improvements.

In Figure~\ref{fig:uniform_speedup}, we demonstrate the speedup of our method over LKH-3, comparing to baseline methods. The speedup is often significant even at low levels of solution quality, and improves with higher solution quality. The speedup is not as meaningful beyond 95\% LKH-3 solution quality, as LKH-3 takes a disproportionate amount of time to attain the last 5\% of improvement.

\subsection{Clustered and Mixed CVRP Distributions}

We further examine the framework's performance on CVRP distributions with clusters of cities, such as studied in~\citep{christofides1979loading, uchoa2017new}. We generate a clustered instance by first sampling $(x, y)$ locations of $n_c$ cluster centroids then sampling the cities around the centroids. The mixed distribution samples 50\% of the cities from a uniform distribution and the rest from around cluster centers. We generate a dataset of instances for every $(N, n_c, m)\in\{500, 1000, 2000\}\times\{3, 5, 7\}\times \{\text{Clustered}, \text{Mixed}\}$ and train a single model on the entire dataset. We evaluate the model on validation and test sets of 10 instances per $(N, n_c, m)$ combination (i.e. 60 instances per $N$). Due to space limitation, we provide more details about the data distribution and full results in Appendix~\ref{appsec:clustered}.

\begin{wraptable}{r}{7.5cm}
  \caption{\textbf{Speedup for $N = 2000$ clustered and mixed CVRPs} at 95\% LKH-3 30k solution quality.}
  \label{sample-table}
  \centering
  \begin{tabular}{lllll}
    \toprule
    \multicolumn{2}{c}{Setting} & $n_c = 3$     & $n_c = 5$ & $n_c = 7$ \\
    \midrule
    Clustered &\makecell{ Ours \\ Random} & \makecell{ 26x \\ 11x} & \makecell{ 18x \\ 7.5x} & \makecell{ 25x \\ 9.0x}   \\
    \hline
    Mixed     &\makecell{ Ours \\ Random} &  \makecell{ 13x \\ 6.6x} & \makecell{ 14x \\ 6.4x} & \makecell{ 14x \\ 7.6x} \\ 
    \bottomrule
  \end{tabular}\label{table:cluster}
\end{wraptable} 

Table~\ref{table:cluster} reports speedups of our method and the Random baseline over LKH-3. Our method sees at least 2x speedup over Random in all settings. We see larger speedups for clustered distributions than for mixed or uniform distributions (Table~\ref{tab:models}).

\begin{figure*}
  \centering
  \includegraphics[width=0.95\linewidth]{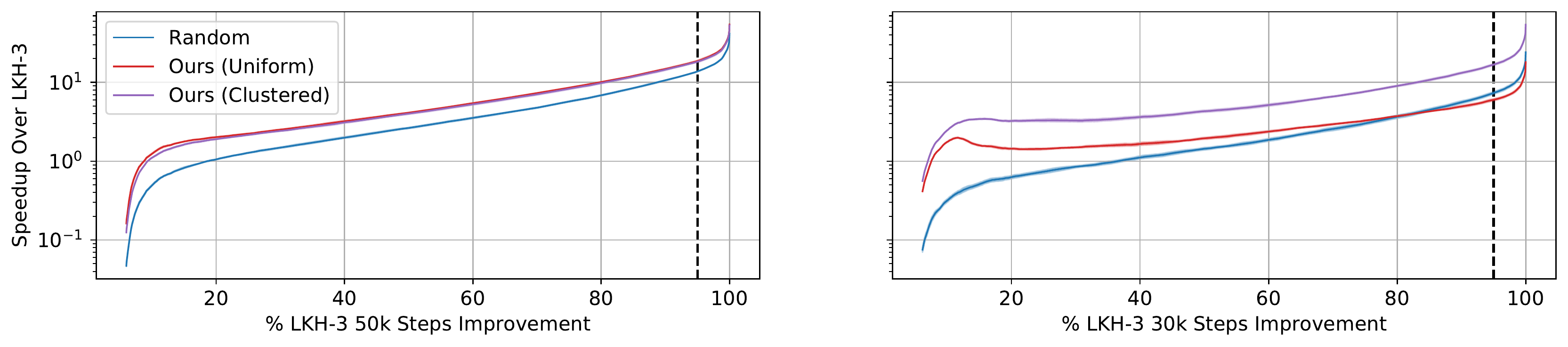}
  \caption{\textbf{Speedup in out-of-distribution CVRPs}. The speedup of our model on the $N = 3000$ uniform distribution (left) and $N = 2000$ real-world distribution (right) without finetuning. Ours (Uniform) and Ours (Clustered) are trained on the uniform and clustered CVRP distributions, respectively. Note that LKH-3 is run for 50k steps for $N = 3000$ instances.}\label{fig:transfer}
\end{figure*}

\subsection{Out-of-distribution Generalization}
\label{sec:generalization}
We study how our subproblem selector generalize to a uniform CVRP distribution with larger problem size $N = 3000$ and to a real-world CVRP distributions, both unseen at training time. The real-world CVRP distribution derives from a CVRP dataset~\citep{arnold2019efficiently} consisting of 10 very-large-scale instances on 5 Belgium regions with $N$ ranging from $3000$ to $30000$ and randomly generated demand distribution. To generate an instance, we subsample the cities in a region without replacement to $N = 2000$, while regenerating the demands to match our training distribution. For each original instance, we generate 5 subsampled instances to form the test set of 50 total instances. We visualize the original and subsampled datasets in Appendix~\ref{appsec:generalization}. 

We apply subproblem selectors trained on uniform and clustered data without finetuning on the new data distributions and report the speedup comparison with the Random baseline in Figure~\ref{fig:transfer}. We see that when transferring to $N=3000$, subproblem selectors trained on uniform and clustered data offer similar performance. However, the model trained on the clustered distribution generalizes well to the real-world distribution while the model trained on the uniform distribution fails to generalize, with worse speedup than the Random baseline. These results suggest that the domain variability of the clustered distribution strongly improves generalization performance.

\subsection{Other VRP Variants}
\label{sec:variants}
While our previous experiments vary the distribution of city locations in CVRP, here we consider two VRP variants with uniform city distribution but with additional constraints: CVRPTW~\citep{solomon1987algorithms} and VRPMPD~\citep{salhi1999cluster}. The former specifies a hard time window constraint for visiting each city, while the latter specifies pick up and delivery constraints in addition to capacity constraint. A detailed description of the variants can be found in Appendix~\ref{appsec:variants}.

Similar to CVRP, we observe significant speedup with our iterative framework alone, while learning offers additional speedup. For CVRPTW, our method offers a 8.2x speedup while our Random baseline offers a 5.9x speedup; for VRPMPD, our method offers a 31x speedup while our Random baseline offers a 20x speedup. We suspect that the time window constraint in CVRPTW imposes strict orderings on the order of city visitations, increasing the difficulty for the subproblem selector.

\begin{figure*}
  \centering
  \includegraphics[width=0.95\linewidth]{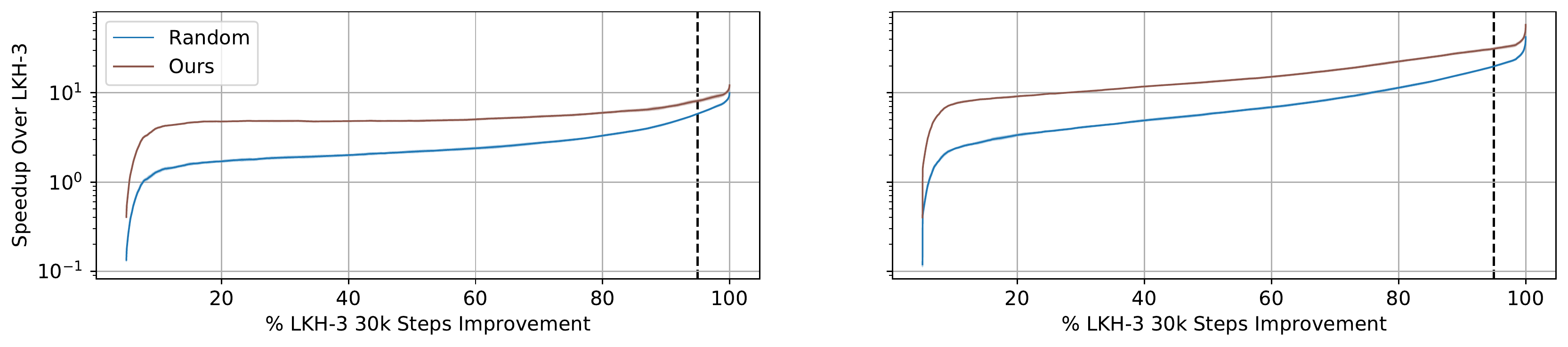}
  \caption{\textbf{Speedup in other $N = 2000$ VRP variants}, CVRPTW (left) and VRPMPD (right).}\label{fig:variants}
\end{figure*}

\subsection{A State-of-the-Art CVRP Subsolver: HGS}
While we focus our analysis on LKH-3 due to its applicability to many variants of VRPs, HGS~\cite{vidal2012hybrid, vidal2020hybrid} is a state-of-the-art solver focused on CVRP. Thus, we apply our method to train subproblem selectors for the HGS subsolver. We discuss the full experimental setup and results in Appendix~\ref{appsec:subsolver}.

With HGS as the subsolver, we observe a 103x speedup for our method on $N = 2000$, compared with a 77x speedup for the Random baseline. Similarly, we observe a 198x speedup for our method on $N = 3000$, compared with a 152x speedup for our Random baseline. The large speedup may be due to the fact that HGS is designed and calibrated for medium-scale problems of 500 to 1000 cities, allowing it to function  better as a subsolver for large-scale VRPs.

\begin{figure*}
  \centering
  \includegraphics[width=0.95\linewidth]{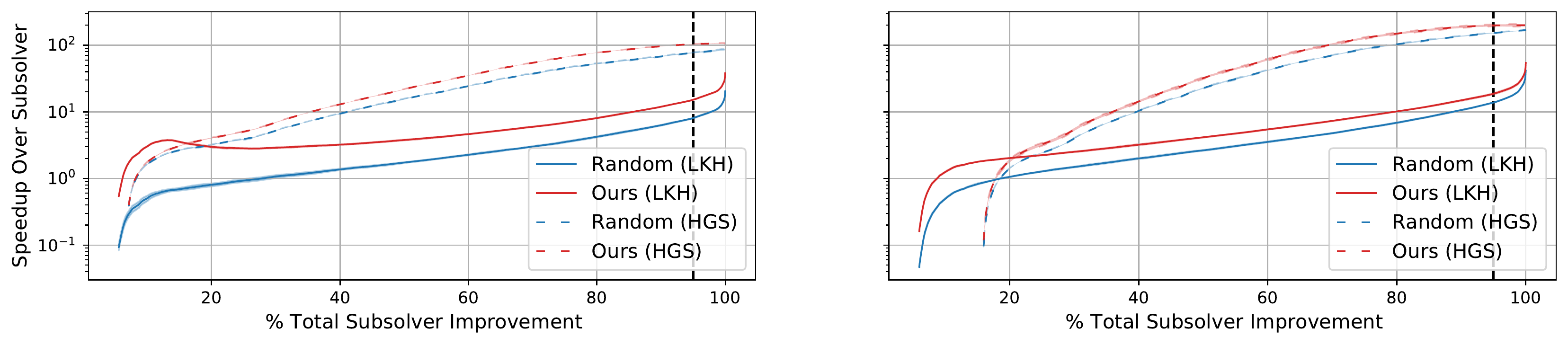}
  \caption{\textbf{Speedup with the HGS subsolver on uniform CVRP}. We compare the speedup of our method equipped with LKH-3 or HGS as the subsolver for $N = 2000$ (left) and $N = 3000$ (right).}\label{fig:variants}
\end{figure*}

\section{Conclusion}
\label{sec:conclusion}
This paper presents a learning-based framework which learns which subproblems to delegate to a subsolver when solving large VRPs. Spatial locality allows us to learn the subproblem selector over a reduced selection space. The proposed method accelerates competitive VRP solvers on problems of sizes up to $3000$, requiring an order of magnitude less computation time. We identify a 1.5x to 2x speedup over non-learning selection strategies. Our results generalize to a variety of VRP distributions, variants, and solvers. 

While most previous learning-based combinatorial optimization methods~\cite{kool2018attention, NEURIPS2020_f231f210, ahn2020learning} rely on reinforcement learning due to the unavailability of optimal solutions as high-quality labels, our work highlights the counterintuitive effectiveness of supervised learning with only \textit{moderate-quality} labels to achieve high-quality solutions with iterative subproblem selection. An interesting line of future work may explore other ways to effectively leverage moderate-quality labels for combinatorial optimization tasks. In particular, we discuss in Appendix~\ref{appsec:applicability} the applicability of our method to other combinatorial optimization problems with spatial locality. We believe that our learning framework can serve as a powerful technique for both the learning and operations research communities to scale up combinatorial optimization solvers.

Our code is publicly available at \url{https://github.com/mit-wu-lab/learning-to-delegate}.

\paragraph{Negative Social Impact.} Enabling more efficient solutions of large-scale VRPs may exhibit negative externalities such as inducing additional traffic from delivery vehicles and centralizing services that pose a stronger competition to brick-and-mortar retail.

\section{Acknowledgements}
%https://neurips.cc/Conferences/2021/PaperInformation/FundingDisclosure
This research was supported by MIT Indonesia Seed Fund, US DOT DDETFP, and the MIT-IBM Watson AI Lab. The authors are grateful to the anonymous reviewers for detailed comments that substantially improved the article. The authors acknowledge the MIT SuperCloud and Lincoln Laboratory Supercomputing Center for providing (HPC, database, consultation) resources that have contributed to the research results reported within this paper. We also thank Zongyi Li for helpful discussions and technical advice throughout the project.

\bibliographystyle{apalike}
\bibliography{ref}

\begin{thebibliography}{}

\bibitem[Aarts et~al., 2003]{aarts2003local}
Aarts, E., Aarts, E.~H., and Lenstra, J.~K. (2003).
\newblock {\em Local search in combinatorial optimization}.
\newblock Princeton University Press.

\bibitem[Ahn et~al., 2020]{ahn2020learning}
Ahn, S., Seo, Y., and Shin, J. (2020).
\newblock Learning what to defer for maximum independent sets.
\newblock In {\em International Conference on Machine Learning}, pages
  134--144. PMLR.

\bibitem[Alvim and Taillard, 2013]{alvim2013popmusic}
Alvim, A.~C. and Taillard, E.~D. (2013).
\newblock Popmusic for the world location-routing problem.
\newblock {\em EURO Journal on Transportation and Logistics}, 2(3):231--254.

\bibitem[Arnold et~al., 2019]{arnold2019efficiently}
Arnold, F., Gendreau, M., and S{\"o}rensen, K. (2019).
\newblock Efficiently solving very large-scale routing problems.
\newblock {\em Computers \& Operations Research}, 107:32--42.

\bibitem[B{\l}a{\.z}ewicz et~al., 1996]{blazewicz1996job}
B{\l}a{\.z}ewicz, J., Domschke, W., and Pesch, E. (1996).
\newblock The job shop scheduling problem: Conventional and new solution
  techniques.
\newblock {\em European journal of operational research}, 93(1):1--33.

\bibitem[Bosman and La~Poutr{\'e}, 2006]{bosman2006computationally}
Bosman, P.~A. and La~Poutr{\'e}, H. (2006).
\newblock Computationally intelligent online dynamic vehicle routing by
  explicit load prediction in an evolutionary algorithm.
\newblock In {\em Parallel Problem Solving from Nature-PPSN IX}, pages
  312--321. Springer.

\bibitem[Chen and Tian, 2019]{NEURIPS2019_131f383b}
Chen, X. and Tian, Y. (2019).
\newblock Learning to perform local rewriting for combinatorial optimization.
\newblock In Wallach, H., Larochelle, H., Beygelzimer, A., d~Alch\'{e}-Buc, F.,
  Fox, E., and Garnett, R., editors, {\em Advances in Neural Information
  Processing Systems}, volume~32. Curran Associates, Inc.

\bibitem[Christofides et~al., 1979]{christofides1979loading}
Christofides, N., Mingozzi, A., and Toth, P. (1979).
\newblock Loading problems.
\newblock {\em N. Christofides and al., editors, Combinatorial Optimization},
  pages 339--369.

\bibitem[Croes, 1958]{croes1958method}
Croes, G.~A. (1958).
\newblock A method for solving traveling-salesman problems.
\newblock {\em Operations research}, 6(6):791--812.

\bibitem[Dorigo et~al., 2006]{dorigo2006ant}
Dorigo, M., Birattari, M., and Stutzle, T. (2006).
\newblock Ant colony optimization.
\newblock {\em IEEE computational intelligence magazine}, 1(4):28--39.

\bibitem[Gehring and Homberger, 1999]{gehring1999parallel}
Gehring, H. and Homberger, J. (1999).
\newblock A parallel hybrid evolutionary metaheuristic for the vehicle routing
  problem with time windows.
\newblock In {\em Proceedings of EUROGEN99}, volume~2, pages 57--64. Citeseer.

\bibitem[Gurobi~Optimization, 2021]{gurobi}
Gurobi~Optimization, L. (2021).
\newblock Gurobi optimizer reference manual.

\bibitem[Helsgaun, 2017]{helsgaun2017}
Helsgaun, K. (2017).
\newblock An extension of the lin-kernighan-helsgaun tsp solver for constrained
  traveling salesman and vehicle routing problems.
\newblock \url{http://akira.ruc.dk/~keld/research/LKH-3/}.

\bibitem[Helsgaun, 2018]{helsgaun2018using}
Helsgaun, K. (2018).
\newblock Using popmusic for candidate set generation in the
  lin-kernighan-helsgaun tsp solver.
\newblock {\em Roskilde Universitet}, 7.

\bibitem[Hottung and Tierney, 2020]{hottung2019neural}
Hottung, A. and Tierney, K. (2020).
\newblock Neural large neighborhood search for the capacitated vehicle routing
  problem.
\newblock In {\em 24th European Conference on Artificial Intelligence (ECAI
  2020)}.

\bibitem[Huber, 1992]{huber1992robust}
Huber, P.~J. (1992).
\newblock Robust estimation of a location parameter.
\newblock In {\em Breakthroughs in statistics}, pages 492--518. Springer.

\bibitem[Kallehauge et~al., 2005]{kallehauge2005vehicle}
Kallehauge, B., Larsen, J., Madsen, O.~B., and Solomon, M.~M. (2005).
\newblock Vehicle routing problem with time windows.
\newblock In {\em Column generation}, pages 67--98. Springer.

\bibitem[Khalil et~al., 2017]{dai2017learning}
Khalil, E.~B., Dai, H., Zhang, Y., Dilkina, B., and Song, L. (2017).
\newblock Learning combinatorial optimization algorithms over graphs.
\newblock In {\em Advances in Neural Information Processing Systems}.

\bibitem[Kohl et~al., 1999]{kohl19992}
Kohl, N., Desrosiers, J., Madsen, O.~B., Solomon, M.~M., and Soumis, F. (1999).
\newblock 2-path cuts for the vehicle routing problem with time windows.
\newblock {\em Transportation Science}, 33(1):101--116.

\bibitem[Kool et~al., 2019]{kool2018attention}
Kool, W., van Hoof, H., and Welling, M. (2019).
\newblock Attention, learn to solve routing problems!
\newblock In {\em International Conference on Learning Representations}.

\bibitem[Kwon et~al., 2020]{NEURIPS2020_f231f210}
Kwon, Y.-D., Choo, J., Kim, B., Yoon, I., Gwon, Y., and Min, S. (2020).
\newblock Pomo: Policy optimization with multiple optima for reinforcement
  learning.
\newblock In Larochelle, H., Ranzato, M., Hadsell, R., Balcan, M.~F., and Lin,
  H., editors, {\em Advances in Neural Information Processing Systems},
  volume~33, pages 21188--21198. Curran Associates, Inc.

\bibitem[Lalla-Ruiz and Voss, 2016]{lalla2016popmusic}
Lalla-Ruiz, E. and Voss, S. (2016).
\newblock Popmusic as a matheuristic for the berth allocation problem.
\newblock {\em Annals of Mathematics and Artificial Intelligence},
  76(1-2):173--189.

\bibitem[Laporte, 2009]{laporte2009fifty}
Laporte, G. (2009).
\newblock Fifty years of vehicle routing.
\newblock {\em Transportation science}, 43(4):408--416.

\bibitem[Laurent et~al., 2009]{laurent2009point}
Laurent, M., Taillard, {\'E}.~D., Ertz, O., Grin, F., Rappo, D., and Roh, S.
  (2009).
\newblock From point feature label placement to map labelling.
\newblock In {\em Proceedings, metaheuristic international conference
  (MIC’09), Hamburg}. Citeseer.

\bibitem[Lenstra and Kan, 1981]{lenstra1981complexity}
Lenstra, J.~K. and Kan, A.~R. (1981).
\newblock Complexity of vehicle routing and scheduling problems.
\newblock {\em Networks}, 11(2):221--227.

\bibitem[Lin and Kernighan, 1973]{lin1973effective}
Lin, S. and Kernighan, B.~W. (1973).
\newblock An effective heuristic algorithm for the traveling-salesman problem.
\newblock {\em Operations research}, 21(2):498--516.

\bibitem[Lu et~al., 2020]{Lu2020A}
Lu, H., Zhang, X., and Yang, S. (2020).
\newblock A learning-based iterative method for solving vehicle routing
  problems.
\newblock In {\em International Conference on Learning Representations}.

\bibitem[Lysgaard et~al., 2004]{lysgaard2004}
Lysgaard, J., Letchford, A.~N., and Eglese, R.~W. (2004).
\newblock A new branch-and-cut algorithm for the capacitated vehicle routing
  problem.
\newblock {\em Mathematical Programming}, 100(2):423--445.

\bibitem[Nazari et~al., 2018]{nazari2018reinforcement}
Nazari, M., Oroojlooy, A., Snyder, L., and Takac, M. (2018).
\newblock Reinforcement learning for solving the vehicle routing problem.
\newblock In Bengio, S., Wallach, H., Larochelle, H., Grauman, K.,
  Cesa-Bianchi, N., and Garnett, R., editors, {\em Advances in Neural
  Information Processing Systems}, volume~31. Curran Associates, Inc.

\bibitem[Perron and Furnon, 2019]{ortools}
Perron, L. and Furnon, V. (2019).
\newblock Or-tools.

\bibitem[Poullet, 2020]{poullet2020leveraging}
Poullet, J. (2020).
\newblock {\em Leveraging machine learning to solve The vehicle Routing Problem
  with Time Windows}.
\newblock PhD thesis, Massachusetts Institute of Technology.

\bibitem[Renaud and Boctor, 2002]{renaud2002sweep}
Renaud, J. and Boctor, F.~F. (2002).
\newblock A sweep-based algorithm for the fleet size and mix vehicle routing
  problem.
\newblock {\em European Journal of Operational Research}, 140(3):618--628.

\bibitem[Salhi and Nagy, 1999]{salhi1999cluster}
Salhi, S. and Nagy, G. (1999).
\newblock A cluster insertion heuristic for single and multiple depot vehicle
  routing problems with backhauling.
\newblock {\em Journal of the operational Research Society}, 50(10):1034--1042.

\bibitem[Santini et~al., 2021]{santini2021decomposition}
Santini, A., Schneider, M., Vidal, T., and Vigo, D. (2021).
\newblock Decomposition strategies for vehicle routing heuristics.

\bibitem[Shaw, 1998]{shaw1998using}
Shaw, P. (1998).
\newblock Using constraint programming and local search methods to solve
  vehicle routing problems.
\newblock In {\em International conference on principles and practice of
  constraint programming}, pages 417--431. Springer.

\bibitem[Sivanandam and Deepa, 2008]{sivanandam2008genetic}
Sivanandam, S. and Deepa, S. (2008).
\newblock Genetic algorithms.
\newblock In {\em Introduction to genetic algorithms}, pages 15--37. Springer.

\bibitem[Solomon, 1987]{solomon1987algorithms}
Solomon, M.~M. (1987).
\newblock Algorithms for the vehicle routing and scheduling problems with time
  window constraints.
\newblock {\em Operations research}, 35(2):254--265.

\bibitem[Song et~al., 2020]{song2020general}
Song, J., Lanka, R., Yue, Y., and Dilkina, B. (2020).
\newblock A general large neighborhood search framework for solving integer
  linear programs.
\newblock In Larochelle, H., Ranzato, M., Hadsell, R., Balcan, M.~F., and Lin,
  H., editors, {\em Advances in Neural Information Processing Systems},
  volume~33, pages 20012--20023. Curran Associates, Inc.

\bibitem[Taillard, 2003]{taillard2003heuristic}
Taillard, {\'E}.~D. (2003).
\newblock Heuristic methods for large centroid clustering problems.
\newblock {\em Journal of heuristics}, 9(1):51--73.

\bibitem[Taillard et~al., 2001]{taillard2001adaptive}
Taillard, {\'E}.~D., Gambardella, L.~M., Gendreau, M., and Potvin, J.-Y.
  (2001).
\newblock Adaptive memory programming: A unified view of metaheuristics.
\newblock {\em European Journal of Operational Research}, 135(1):1--16.

\bibitem[Taillard and Helsgaun, 2019]{taillard2019popmusic}
Taillard, {\'E}.~D. and Helsgaun, K. (2019).
\newblock Popmusic for the travelling salesman problem.
\newblock {\em European Journal of Operational Research}, 272(2):420--429.

\bibitem[Taillard and Voss, 2002]{taillard2002popmusic}
Taillard, {\'E}.~D. and Voss, S. (2002).
\newblock Popmusic—partial optimization metaheuristic under special
  intensification conditions.
\newblock In {\em Essays and surveys in metaheuristics}, pages 613--629.
  Springer.

\bibitem[Taillard and Vo{\ss}, 2018]{Taillard2018}
Taillard, {\'E}.~D. and Vo{\ss}, S. (2018).
\newblock {\em POPMUSIC}, pages 687--701.
\newblock Springer International Publishing, Cham.

\bibitem[Uchoa et~al., 2017]{uchoa2017new}
Uchoa, E., Pecin, D., Pessoa, A., Poggi, M., Vidal, T., and Subramanian, A.
  (2017).
\newblock New benchmark instances for the capacitated vehicle routing problem.
\newblock {\em European Journal of Operational Research}, 257(3):845--858.

\bibitem[Vaswani et~al., 2017]{vaswani2017attention}
Vaswani, A., Shazeer, N., Parmar, N., Uszkoreit, J., Jones, L., Gomez, A.~N.,
  Kaiser, L.~u., and Polosukhin, I. (2017).
\newblock Attention is all you need.
\newblock In Guyon, I., Luxburg, U.~V., Bengio, S., Wallach, H., Fergus, R.,
  Vishwanathan, S., and Garnett, R., editors, {\em Advances in Neural
  Information Processing Systems}, volume~30. Curran Associates, Inc.

\bibitem[Veli{\v{c}}kovi{\'{c}} et~al., 2018]{velickovic2018graph}
Veli{\v{c}}kovi{\'{c}}, P., Cucurull, G., Casanova, A., Romero, A., Li{\`{o}},
  P., and Bengio, Y. (2018).
\newblock {Graph Attention Networks}.
\newblock {\em International Conference on Learning Representations}.

\bibitem[Ventresca et~al., 2013]{ventresca2013predicting}
Ventresca, M., Ombuki-Berman, B., and Runka, A. (2013).
\newblock Predicting genetic algorithm performance on the vehicle routing
  problem using information theoretic landscape measures.
\newblock In {\em European Conference on Evolutionary Computation in
  Combinatorial Optimization}, pages 214--225. Springer.

\bibitem[Vidal, 2020]{vidal2020hybrid}
Vidal, T. (2020).
\newblock Hybrid genetic search for the cvrp: Open-source implementation and
  swap* neighborhood.
\newblock {\em arXiv preprint arXiv:2012.10384}.

\bibitem[Vidal et~al., 2012]{vidal2012hybrid}
Vidal, T., Crainic, T.~G., Gendreau, M., Lahrichi, N., and Rei, W. (2012).
\newblock A hybrid genetic algorithm for multidepot and periodic vehicle
  routing problems.
\newblock {\em Operations Research}, 60(3):611--624.

\bibitem[Vinyals et~al., 2015]{vinyals2015pointer}
Vinyals, O., Fortunato, M., and Jaitly, N. (2015).
\newblock Pointer networks.
\newblock In Cortes, C., Lawrence, N., Lee, D., Sugiyama, M., and Garnett, R.,
  editors, {\em Advances in Neural Information Processing Systems}, volume~28.
  Curran Associates, Inc.

\bibitem[Voudouris and Tsang, 2003]{voudouris2003guided}
Voudouris, C. and Tsang, E.~P. (2003).
\newblock Guided local search.
\newblock In {\em Handbook of metaheuristics}, pages 185--218. Springer.

\bibitem[Wu et~al., 2016]{wu2016optimizing}
Wu, C., Shankari, K., Kamar, E., Katz, R., Culler, D., Papadimitriou, C.,
  Horvitz, E., and Bayen, A. (2016).
\newblock Optimizing the diamond lane: A more tractable carpool problem and
  algorithms.
\newblock In {\em 2016 IEEE 19th International Conference on Intelligent
  Transportation Systems (ITSC)}, pages 1389--1396. IEEE.

\bibitem[Yolcu and Poczos, 2019]{NEURIPS2019_12e59a33}
Yolcu, E. and Poczos, B. (2019).
\newblock Learning local search heuristics for boolean satisfiability.
\newblock In Wallach, H., Larochelle, H., Beygelzimer, A., d\textquotesingle
  Alch\'{e}-Buc, F., Fox, E., and Garnett, R., editors, {\em Advances in Neural
  Information Processing Systems}, volume~32. Curran Associates, Inc.

\end{thebibliography}

%%%%%%%%%%%%%%%%%%%%%%%%%%%%%%%%%%%%%%%%%%%%%%%%%%%%%%%%%%%%
\newpage
\section*{Checklist}
\begin{enumerate}

\item For all authors...
\begin{enumerate}
  \item Do the main claims made in the abstract and introduction accurately reflect the paper's contributions and scope?
    \answerYes{}
  \item Did you describe the limitations of your work?
    \answerYes{See Section~\ref{sec:variants}. We discuss our method gets slightly less speedup on CVRPTW than CVRP or VRPMPD; in section~\ref{sec:uniform}, we also discuss our learned subproblem-selection order at convergence are the same as Random and Min Count order, which can be seen as a potential limitation, but we provide justification in the same Section (\ref{sec:uniform}) as well.}
  \item Did you discuss any potential negative societal impacts of your work?
    \answerYes{See the negative social impact paragraph in Section~\ref{sec:conclusion} (Conclusion)}
  \item Have you read the ethics review guidelines and ensured that your paper conforms to them?
    \answerYes{}
\end{enumerate}

\item If you are including theoretical results...
\begin{enumerate}
  \item Did you state the full set of assumptions of all theoretical results?
    \answerNA{}
	\item Did you include complete proofs of all theoretical results?
    \answerNA{}
\end{enumerate}

\item If you ran experiments...
\begin{enumerate}
  \item Did you include the code, data, and instructions needed to reproduce the main experimental results (either in the supplemental material or as a URL)?
    \answerYes{As stated in Section~\ref{sec:conclusion} (Conclusion), our code is released at \url{https://github.com/mit-wu-lab/learning-to-delegate}.}
  \item Did you specify all the training details (e.g., data splits, hyperparameters, how they were chosen)?
    \answerYes{We briefly mention the training details in Section~\ref{sec:setup} of the main paper, and we put the full details in Appendix~\ref{appsec:uniform_setup}.}
	\item Did you report error bars (e.g., with respect to the random seed after running experiments multiple times)?
    \answerYes{We plot error bars on all of our experiment figures in Section~\ref{sec:experiment}.}
	\item Did you include the total amount of compute and the type of resources used (e.g., type of GPUs, internal cluster, or cloud provider)?
    \answerYes{See Section~\ref{sec:setup}.}
\end{enumerate}

\item If you are using existing assets (e.g., code, data, models) or curating/releasing new assets...
\begin{enumerate}
  \item If your work uses existing assets, did you cite the creators?
    \answerYes{We use the LKH-3 and HGS VRP solvers as a part of our component. We cite the creator as~\citep{helsgaun2017} (LKH-3) and~\citep{vidal2012hybrid, vidal2020hybrid} (HGS) in our paper. We compare with existing learning and heuristic baseline using their code. We cite the creators in our paper, and provide exerpeiment setup (including github links to the repos) in the supplementary material.}
  \item Did you mention the license of the assets?
    \answerNA{All the assets (code and data) we use are open source.}
  \item Did you include any new assets either in the supplemental material or as a URL?
    \answerNo{}
  \item Did you discuss whether and how consent was obtained from people whose data you're using/curating?
    \answerNA{We generate synthetic data for most of our experiments. The only real dataset in Section~\ref{sec:generalization} is open source.}
  \item Did you discuss whether the data you are using/curating contains personally identifiable information or offensive content?
    \answerNA{}
\end{enumerate}

\item If you used crowdsourcing or conducted research with human subjects...
\begin{enumerate}
  \item Did you include the full text of instructions given to participants and screenshots, if applicable?
    \answerNA{}
  \item Did you describe any potential participant risks, with links to Institutional Review Board (IRB) approvals, if applicable?
    \answerNA{}
  \item Did you include the estimated hourly wage paid to participants and the total amount spent on participant compensation?
    \answerNA{}
\end{enumerate}

\end{enumerate}

%%%%%%%%%%%%%%%%%%%%%%%%%%%%%%%%%%%%%%%%%%%%%%%%%%%%%%%%%%%%

\newpage
\appendix
\section{Appendix}
\label{app}

\localtableofcontents

% \newpage
\subsection{Experiment Setup}
\label{appsec:uniform_setup}
We discuss the full experimental setup for training and evaluating the subproblem selector on the uniform CVRP distribution in this section while reserving any minor modifications for the clustered CVRP distributions, real-world CVRP distribution, VRP variants, and ablation studies for Appendix~\ref{appsec:clustered}, \ref{appsec:generalization}, \ref{appsec:variants}, and \ref{appsec:ablation}, respectively.

\paragraph{Uniform CVRP distribution.} We generate CVRP instances of size $N=500, 1000, 2000,$ and $3000$ following the same distribution of cities locations and demands as previous works~\citep{nazari2018reinforcement, kool2018attention, Lu2020A, NEURIPS2020_f231f210}: the depot and cities' $(x,y)$ locations are sampled uniformly from $[0, 1]^2$, each city's demand $d$ is sampled uniformly from $\{1, 2, ..., 9\}$, and each vehicle has capacity $C=50$. For each $N$, we sample $n_\text{train} = 2000$, $n_\text{val} = 40$, and $n_\text{test} = 40$ problem instances for the training, validation, and test set.

\begin{table}[ht]
\begin{minipage}{.5\linewidth}
\caption{\textbf{Uniform CVRP parameters.}}
\centering
\begin{tabular}{cc}
\toprule % from booktabs package
Choices of $N$ & $\{500, 1000, 2000, 3000\}$\\
Dist. of depot location & $\mathcal U([0, 1]^2)$\\
Dist. of city location & $\mathcal U([0, 1]^2)$\\
Dist. of city demand & $\mathcal U(\{1, \dots 9\})$\\
Vehicle capacity & $50$\\
\bottomrule % from booktabs package
\end{tabular}
\end{minipage}%
\begin{minipage}{.5\linewidth}
\centering
\caption{\textbf{Data splits.}}
\begin{tabular}{cc}
\toprule % from booktabs package
Training & 2000 instances\\
Validation & 40 instances\\
Test & 40 instances\\
\bottomrule % from booktabs package
\end{tabular}
\end{minipage} 
\end{table}

\paragraph{Initialization solution.} Our iterative learning-to-delegate framework (Figure~\ref{fig:pipeline}) requires a feasible initial solution $X_0$. To generate $X_0$ for each problem instance, we employ a fixed initialization scheme in which the space is partitioned into $10$ equally-sized angular sectors radiating outward from the depot. We then run the LKH-3 solver for a brief $100$ steps on each partition to produce initial routes. Initialization on average takes 6 seconds for $N=500$, 18 seconds for $N=1000$, 50 seconds for $N=2000$, and 94 seconds for $N=3000$ on a single CPU. Our initialization scheme is similar to the sweep-based algorithm proposed by \citet{renaud2002sweep}, which is a commonly used initialization scheme for iterative VRP solvers. Unless stated otherwise, we use the same initialization scheme in all methods to fairly compare each method's ability to improve from a rough initialization. We additionally explore the effect of initialization quality on our method in Appendix~\ref{appsec:initialization}.

\paragraph{Training data generation.} We generate the training set by running our iterative framework and selecting subproblems as follows: for each instance $P$ with initial solution $X$ in the training set, we run the subsolver on each $S \in \mathcal S_{k, \text{local}}$ to compute the subsolution $X_S$ and improvement $\delta(S)$. We update the solution from $X$ to $X'$ with $\arg\max_{S} \delta(S)$ then repeat the process on $X'$ for a fixed number of steps $D_\text{train} = 30$ of our iterative framework. For the uniform CVRP distribution, for each route $r$ in the current solution $X$, we construct $\mathcal S_{k, \text{local}}$ with $k = 10$ nearest routes. We compute the subsolution by running the LKH-3 subsolver for 500 steps, which takes around 6.7 seconds on a single CPU for a $k = 10$ subproblem composed of around $100$ cities. We concatenate training data from multiple $N$'s to train the subproblem regression model. As mentioned in the main text, most subproblems remain unchanged from $X$ to $X'$, so we do not repeat unchanged subproblems in our generated training set; therefore, the number of unique subproblems in our training set may be around 3-8 times smaller than the total number of non-unique subproblems $n_\text{train} D_\text{train} \mathbb E[R] \approx n_\text{train} D_\text{train} \frac{N}{\mathbb E[d]}$ in the training set, where $\mathbb E[R]$ is the average number of routes in a solution.

As we find that the performance of the subproblem regression model trained on the concatenated $N = 500$ and $1000$ data offers satisfactory performance, even when transferred to $N = 2000$ and $3000$, we do not collect training data for the latter two cases. Collecting training data is the most computationally intensive component of our work, and the computation time increases with larger $N$, larger $k$, and larger $D_\text{train}$. By restricting training data collection to $N = 500$ and $1000$, $k = 10$, and $D_\text{train} = 30$, the total time taken is around 10 hours on 200 CPUs.

\paragraph{Input features.} To convert each subproblem $S$ into the input to our Transformer subproblem selector, we create a $3$-dim input feature vector for each city consisting of the city's $(x, y)$ locations (centered around the depot's $(x, y)$ location) and the demand $d$ (rescaled by the capacity).

\begin{table}
\begin{minipage}{.5\linewidth}
\caption{\textbf{Best architecture hyperparameters.}}
\centering
\begin{tabular}{cc}
\toprule % from booktabs package
Input dimension & 3\\
Base model & Transformer\\
Model dimension $d_\text{model}$ & 128\\
Number of heads $n_\text{heads}$ & 8\\
Number of layers $n_\text{layers}$ & 6\\
Feed-forward dimension $d_\text{ff}$ & 512\\
Activation & ReLU\\
Layer Normalization & Yes\\
Dropout & 0\\
\bottomrule % from booktabs package
\end{tabular}
\end{minipage}%
\begin{minipage}{.5\linewidth}
\centering
\caption{\textbf{Best training hyperparameters.}}
\begin{tabular}{cc}
\toprule % from booktabs package
Optimizer & Adam\\
Learning rate & 0.001\\
Learning rate schedule & Cosine\\
Batch size & 512\\
Number of gradient steps & 40000\\
GPU & NVIDIA V100\\
\bottomrule
\end{tabular}
\end{minipage} 
\end{table}

\paragraph{Architecture and training hyperparameters.} Our best Transformer model uses model dimension $d_\text{model} = 128$, $n_\text{heads} = 8$ heads, $n_\text{layers} = 6$ layers, feed-forward dimension of $d_\text{ff} = 512$, ReLU activation, layer normalization, and no dropout. We train with Adam optimizer with a learning rate of 0.001, cosine learning rate decay, and a batch size of 2048 for 40000 gradient steps. All hyperparameters are selected on the validation set and frozen before evaluating on the test set.

\paragraph{Baseline learning methods.} In Table~\ref{tab:models}, we compare our methods with two learning baselines: Attention Model (AM)~\citep{kool2018attention} and NeuRewriter~\citep{NEURIPS2019_131f383b}. We use the pre-trained AM model provided by the author: \url{https://github.com/wouterkool/attention-learn-to-route}. The model is trained on problems of size 100 following the same distribution as ours. The model has two modes of prediction: greedy and sampling. We perform 1280 samples per instance, yet find the sampling performance worse than the greedy result. While the original paper reports better sampling performance than greedy when $N \leq 100$, we believe that AM sampling suffers from the fact that the solution space is exponentially larger with our large problem instance sizes. Therefore, any small stochasticity and imperfections from AM sampling may autoregressively compound much more in our setting than in previous AM settings. We use a single NVIDIA V100 GPU with 20 CPUs for evaluation.

For NeuRewriter, we re-train the network on problems of size 100 using the code provided by the author: \url{https://github.com/facebookresearch/neural-rewriter}, as no pre-trained model is available. Similarly to AM, we choose size 100 since it is the largest training problem size from the paper. We run into memory and fitting issue when trying to rerun their code on larger problem sizes, so we do not report the numbers here. We use a single NVIDIA V100 GPU with 20 CPUs to perform the evaluation.

\subsection{Clustered and Mixed CVRP}
\label{appsec:clustered}
\paragraph{Clustered CVRP distribution.} We generate clustered CVRP distribution by modifying the distribution of city locations within the uniform CVRP distribution specified in Appendix~\ref{appsec:uniform_setup}. Given the number of clusters $n_c$, we first generate $n_c$ cluster centroids by sampling from $\mathcal U([0.2, 0.8]^2)$ as the $(x, y)$ locations. Then, we generate the city nodes by first sampling a centroid uniformly at random, and then sampling the $(x, y)$ location normally distributed with the mean at the centroid and standard deviation $0.07$. The $(x, y)$ locations are clipped within the $[0, 1]^2$ box.

\paragraph{Mixed CVRP distribution.} For the mixed distributions, we sample half of the city nodes according to the uniform distribution, and the other half according to a clustered distribution with $n_c$ centers. This is the common practice used in standard CVRP benchmark datasets~\citep{uchoa2017new}. Note that we generate our new dataset instead of using the benchmark datasets because the CVRP benchmark datasets include mostly small scale $N \leq 500$ instances and have very few instances with $N \geq 100$.

\paragraph{Dataset composition.} As we experiment over all combinations of $N \in \{500, 1000, 2000\}$, $n_c \in \{3, 5, 7\}$, and $m \in \{\text{Clustered}, \text{Mixed}\}$, we generate smaller training, validation, and test set sizes with 500, 10, and 10 instances respectively for each combination. We visualize one clustered and one mixed instance in Figure~\ref{fig:dists}.

\paragraph{Training and evaluation.} Like in uniform CVRP, we do not collect the training set for the $N = 2000$ instances and concatenate the training set for all remaining combinations to create a single training set of $2 * 3 * 2 * 500 = 6000$ problem instances. We use identical architecture and training hyperparameters as listed in Appendix~\ref{appsec:uniform_setup}, as we find that these hyperparameters perform reasonably well. We report results on the test set of 10 instances for each of the $3 * 3 * 2 = 18$ settings separately.

\paragraph{Additional results.} We include the full evaluation results on the test set for all clustered and mixed distributions. Figure~\ref{fig:cluster_n500} contains results for $N = 500$, Figure~\ref{fig:cluster_n1000} contains results for $N = 1000$, and Figure~\ref{fig:cluster_n2000} contains results for $N = 2000$. As mentioned in the above paragraph, a single trained model for our method is evaluated on all combinations of $(N, n_c, m)$, which demonstrates superior speedup and improvement (when run for a reasonable time period) over the Random baseline. To save space, we also include results with the HGS subsolver in the same set of plots.

\subsection{Out-of-distribution CVRPs}
\label{appsec:generalization}
Here we describe the CVRP distributions and full results supporting the out-of-distribution generalization performance of our trained uniform CVRP and clustered CVRP models, with full training specifications in Appendix~\ref{appsec:uniform_setup} and Appendix~\ref{appsec:clustered} respectively.

\subsubsection{Uniform CVRP with $N = 3000$}
We do not describe the setup here as it was previously described with the uniform CVRP distribution in Appendix~\ref{appsec:uniform_setup}. In Figure~\ref{fig:transfer_n3000}, we see that the model trained on the clustered CVRP demonstrates slightly better improvement than the model trained on the uniform CVRP; otherwise the model performances are similar.

\begin{figure}[ht]
  \centering
  \includegraphics[width=\linewidth]{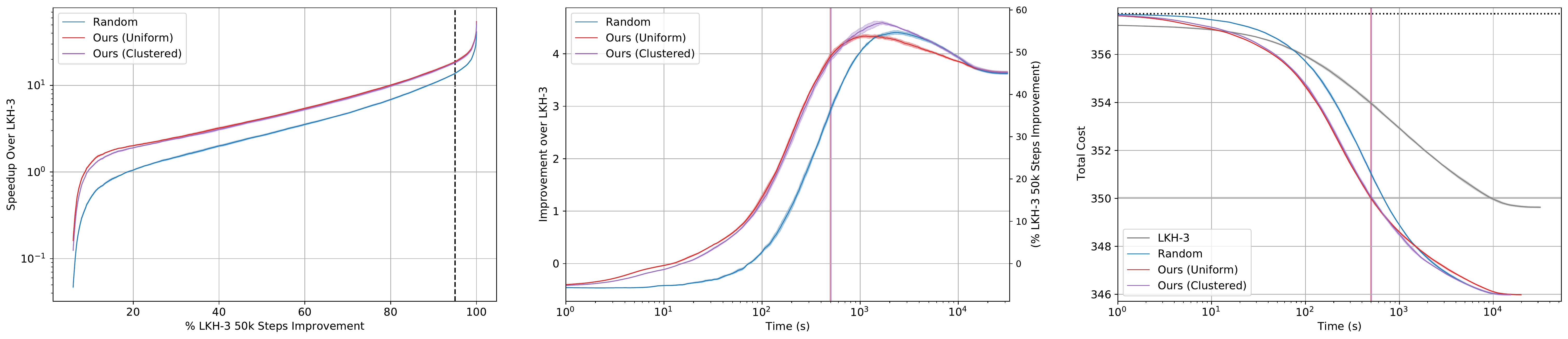}
  \caption{\textbf{Uniform CVRP distributions, $N = 3000$.} Graphs of speedup (left, higher is better), improvement (middle, higher is better), and total cost (right, lower is better). The left graph is the same as the left graph in Figure~\ref{fig:transfer}. The right vertical axis in the improvement graph (middle) indicates the improvement over LKH-3 as a percentage of the total improvement that LKH-3 attains over 30k steps. The horizontal dotted black line in the total cost graph (right) indicates the initial solution quality for all methods. The vertical dashed black line in the speedup graph (left) and the horizontal LKH-3-colored line in the total cost graph (right) indicate the 95\% LKH-3 solution quality. The overlapping colored vertical lines in the improvement graph (middle) and the total cost graph (right) indicate computation times required for the corresponding learning-based methods to reach the aforementioned solution quality.}\label{fig:transfer_n3000}
\end{figure}

\subsubsection{Real-world CVRP}
Excitingly, we see robust generalization performance of our previously trained models to problem instances from an unseen real-world CVRP distribution derived from the large-scale real-world CVRP dataset generated by~\citet{arnold2019efficiently}, which is open source at~\url{https://antor.uantwerpen.be/xxlrouting/}.

\begin{table}[ht]
\centering
\caption{\textbf{Instance sizes $N$ of original real-world instances from \citet{arnold2019efficiently}.} Instances are subsampled without replacement to $N = 2000$ before feeding into our model.}
\label{tab:belgium}
\begin{tabular}{ccc}
\toprule % from booktabs package
\bfseries Region & \bfseries Centralized Depot & \bfseries Eccentric Depot\\
\midrule % from booktabs package
Leuven & 3000 & 4000\\
Antwerp & 6000 & 7000\\
Ghent & 10000 & 11000\\
Brussels & 15000 & 16000\\
Flanders & 20000 & 30000\\
\bottomrule
\end{tabular}
\end{table}

\begin{figure}
  \centering
  \includegraphics[width=\linewidth]{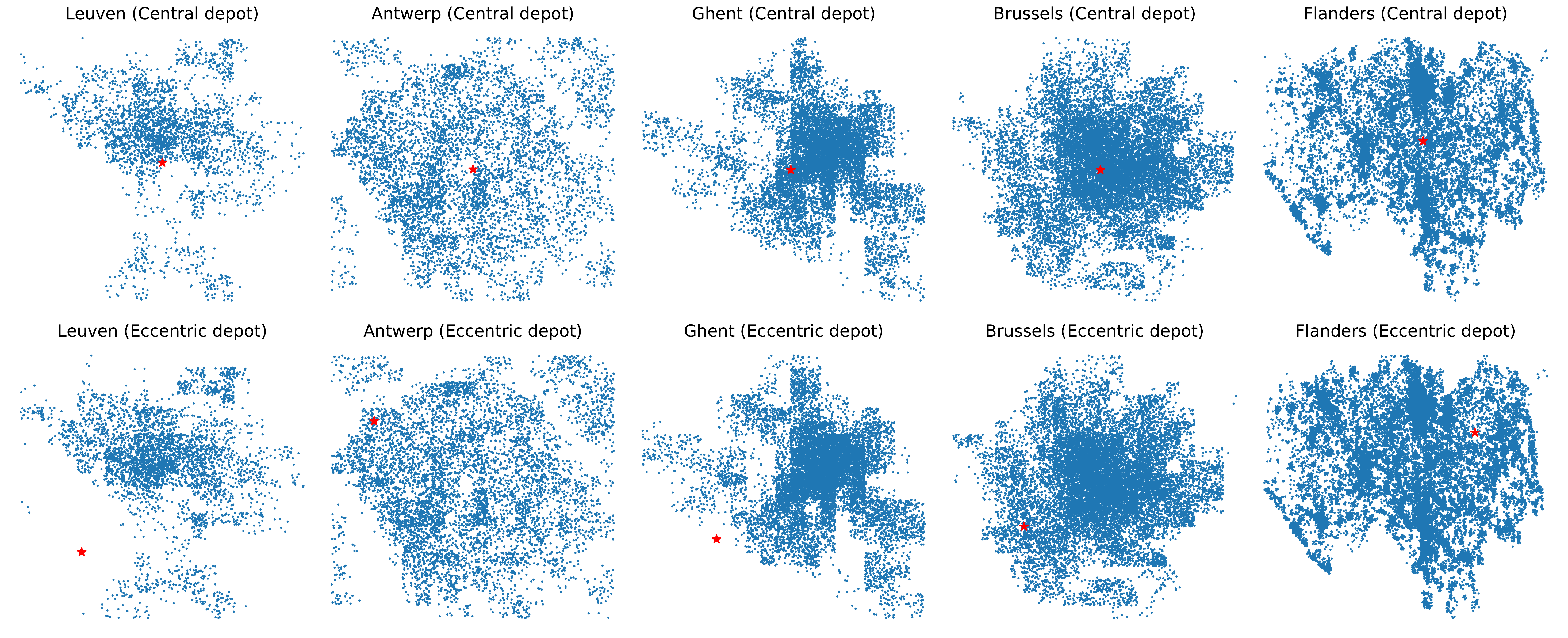}
  \caption{\textbf{Original real-world CVRP instances.} Five instances (top row) have centralized depots and five (bottom row) have eccentric depots.}\label{fig:belgium}
\end{figure}

\paragraph{Original real-world CVRP distribution.}
The original dataset contains 10 instances representing 5 different Belgium regions: Leuven, Antwerp, Ghent, Brussels, and Flanders. For each region, two instances are provided: one with a depot located at the center of all cities and the other with an eccentric depot located in the outskirts of the regions. We list the original sizes of each instance in Table~\ref{tab:belgium} and visualize them in Figure~\ref{fig:belgium}.

This dataset focuses on long-haul deliveries, with the average solution route length ranging from 14.8 to 117.2. The authors randomly sample demand for each node from $\{1, 2, 3\}$ with probabilities 0.5, 0.3, and 0.2 respectively, and the capacity of the vehicle ranges from 20 to 50 for central depot and ranges from 100 to 200 for eccentric depot.

\paragraph{Our real-world CVRP dataset.} To apply our pre-trained models, we first convert the original real-world CVRP to be reasonably close to our training distributions. To generate each instance in our real-world CVRP distribution from an instance in the original dataset, we subsample the original instance to $N = 2000$ without replacement, resample each demand from our demand distribution $\mathcal U(\{1, 2, \dots, 9\})$, then set the vehicle capacity to $C = 50$. For each original instance, we generate 5 instances from our real-world CVRP distribution. Therefore, our validation and test sets contain 50 instances in total. One instance from our test set is visualized in Figure~\ref{fig:dists}.

\begin{figure}[ht]
  \centering
  \includegraphics[width=\linewidth]{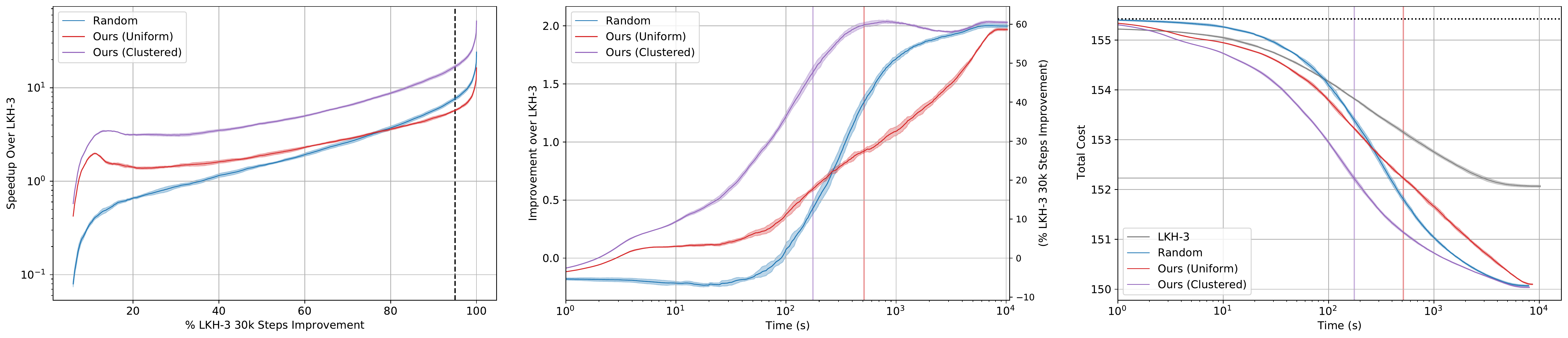}
  \caption{\textbf{Our real-world CVRP distribution.} Graphs of speedup (left, higher is better), improvement (middle, higher is better), and total cost (right, lower is better). The right vertical axis in the improvement graph (middle) indicates the improvement over LKH-3 as a percentage of the total improvement that LKH-3 attains over 30k steps. The horizontal dotted black line in the total cost graph (right) indicates the initial solution quality for all methods. The vertical dashed black line in the speedup graph (left) and the horizontal LKH-3-colored line in the total cost graph (right) indicate the 95\% LKH-3 solution quality. The colored vertical lines in the improvement graph (middle) and the total cost graph (right) indicate computation times required for the corresponding learning-based methods to reach the aforementioned solution quality.}\label{fig:transfer_real}
\end{figure}

\paragraph{Additional results.} We show the full transfer performance of the models trained on uniform and clustered CVRP to our real-world CVRP distribution in Figure~\ref{fig:transfer_real}. We see that the model trained on the clustered CVRP distributions significantly outperforms both the model trained on the uniform CVRP distribution and the Random baseline, whereas the model trained on the uniform CVRP distribution sometimes performs worse than the Random baseline. Encouragingly, the clustered model shows strong generalization performance in CVRP distributions which may be of practical value in the real world.

\subsection{VRP Variants}
\label{appsec:variants}
Here we provide the full definition, experimental setup, and results for VRP variants discussed in Section~\ref{sec:variants}: CVRPTW and VRPMPD.

\subsubsection{CVRPTW (Capacitated Vehicle Routing Problem with Time Windows)}
\label{app:cvrptw}
\paragraph{Problem formulation.} We use the standard CVRPTW formulation found in~\citet{gehring1999parallel} and~\citet{solomon1987algorithms}. In additional to all CVRP constraints, each city node $i$ has a time window $[e_i, l_i]$, the earliest and latest time to visit city $i$, respectively, and a service time $s_i$. The depot has a time window $[e_0, l_0]$ but no service time. Each vehicle departs the depot at time $e_0$ and follows a route $r$ to satisfy the additional time window constraint. Let $i$ and $j$ be consecutive cities on route $r$, and let $b_i^r$ denote the arrival time at city $i$ of the vehicle serving route $r$. Then the departure time at city $i$ is $b_i^r + s_i$ and the arrival time at the next city $j$ is $b_j^r = \max\{e_j, b_i^r + s_i + t_{ij}\}$ where $t_{ij}$ is the travel time that equals the Euclidean distance from $i$ to $j$. This specifies that if a vehicle arrives too early at $j$, then it needs to wait until the start time $e_j$ of $j$. Meanwhile, the route becomes infeasible if the vehicle arrives at $j$ later than $l_j$. Finally, the depot's time window requires the route to travel back to the depot before $l_0$. The number of vehicles is unconstrained. The objective is the same as in CVRP, i.e. minimizing the total Euclidean edge costs of all the routes, which is the standard objective used in the literature~\citep{kohl19992, kallehauge2005vehicle}. The time window constraint imposed by this variant is related to other combinatorial optimization problems such as job shop scheduling~\cite{blazewicz1996job}.

\paragraph{Problem distribution.} We generate CVRPTW instances following a similar procedure as in~\citet{solomon1987algorithms}: we first generate CVRP instances following the same uniform location and demand distribution as described in Section~\ref{appsec:uniform_setup}. For the time window constraint, we set the time window for the depot as $[e_0, l_0] = [0, 3]$, and the service time at each $i$ to be $s_i = 0.2$. We further set the time window for city node $i$ by (1) sampling the time window center $c_i \sim \mathcal U([e_0 + t_{0, i}, l_0 - t_{i, 0} - s_i])$, where $t_{0, i} = t_{i, 0}$ is the travel time, equaling the Euclidean distance, from the depot to node $i$; (2) sampling the time window half-width $h_i$ uniformly at random from $[s_i / 2, l_0 / 3] = [0.1, 1]$; (3) setting the time window for $i$ as $[\max(e_0, c_i - h_i), \min(l_0, c_i + h_i)]$.

\paragraph{Training data generation.} Unlike for CVRP distributions, we generate training data using $k = 5$ routes per subproblem instead of $k = 10$; doing so does not change the number of subproblems at each step, but significantly speeds up the generation process for two reasons: 1) running the subsolver on each subproblem takes around 3.4 seconds instead of 6.7 seconds for $k = 10$ and 2) updating current $X$ with the subsolution $X_S$ for a smaller subproblem $S$ means that there will be more repeated subproblems from $X$ to $X'$ as other subproblems are less likely to be affected. Therefore, in around 10 hours total on 200 CPUs, we are able to generate training data with 2000 instances per $N$ and $D_\text{train} = 40, 80,$ and $160$ for $N = 500, 1000,$ and $2000$, respectively. An ablation study on the effect of $k$ in uniform CVRP is provided in Appendix~\ref{appsec:k5k10}.

\paragraph{Additional results.} We show the full CVRPTW results in Figure~\ref{fig:cvrptw_reg_cls}, which also includes a comparison between subproblem regression and subproblem classification. Interestingly, while our method offers significant speedup over baseline methods and offers significant improvement over LKH-3 (relative to LKH-3's total improvement over 30k steps) when run for a reasonable amount of time, the improvement over the LKH-3 baseline diminishes when run for a very long time. Indeed, our method terminates at a slightly worse total cost than LKH-3 for $N = 500$ when run for a very long time. However, as we show in the ablation over subproblem size $k$ in Appendix~\ref{appsec:k5k10}, we see that using $k = 10$ tends to offer better eventual improvement in uniform CVRP, so the diminishing improvement effect may be partially attributed to the fact that we use $k = 5$ for the CVRPTW experiments.

\subsubsection{VRPMPD (Vehicle Routing Problem with Mixed Pickup and Delivery)}

\paragraph{Problem formulation.} This VRP variant models the backhauling problem where the vehicles are both delivering items to cities from the depot and picking up items from cities to bring back to the depot. We use the standard VRPMPD formulation as in~\citet{salhi1999cluster}, where each city $i$ either has a pickup order with load $p_i$ or delivery order with load $d_i$. Each vehicle with a capacity $C$ starts by loading all delivery orders along the route. Each time the vehicle visits a delivery city $i$, the vehicle's load decreases by $d_i$; conversely, each time the vehicle visits a pickup city $j$, the vehicle's load increases by $p_j$. A valid route requires that the vehicle's load is no greater than $C$ at every point along the route. Therefore, this variant imposes a more complicated capacity constraint than CVRP as the order of city visitation affects the feasibility of a route.

\paragraph{Problem distribution.} We generate VRPMPD instances following a similar procedure as in~\citet{salhi1999cluster}: we take the CVRP instances following the same uniform location and demand distribution as defined in Appendix~\ref{appsec:uniform_setup}, and we randomly assign half the cities pickup orders with $p_i \sim \mathcal U(\{1, 2, \dots, 9\})$ and the other half delivery orders with $d_i \sim \mathcal U(\{1, 2, \dots, 9\})$. We set the vehicle's capacity to be $C = 25$ instead of $50$ (as done for other VRPs) because each route in VRPMPD visits roughly 2x more cities than those in CVRP or CVRPTW with the same capacity, as the vehicle gains empty space after visiting the cities with delivery orders and can serve cities with pickup orders afterward.

\paragraph{Training data generation.} We use the same $k = 5$ data generation process as described in Appendix~\ref{app:cvrptw}, but for VRPMPD instead of CVRPTW.

\paragraph{Additional results.} We show the full VRPMPD results in Figure~\ref{fig:vrpmpd_reg_cls}, which also includes a comparison between subproblem regression and subproblem classification. Similar to the behavior in CVRPTW, while our method offers significant speedup over baseline methods and offers significant improvement over LKH-3 (relative to LKH-3's total improvement over 30k steps) when run for a reasonable amount of time, the improvement over the LKH-3 baseline diminishes when run for a very long time. However, as we show in the ablation over subproblem size $k$ in Appendix~\ref{appsec:k5k10}, we see that using $k = 10$ tends to offer better eventual improvement in uniform CVRP, so the diminishing improvement effect may be partially attributed to the fact that we use $k = 5$ for the VRPMPD experiments.

\subsection{Ablation Studies}
\label{appsec:ablation}

\subsubsection{Regression vs Classification for Subproblem Selection}
\label{appsec:classification}
While we focus on regression in Section~\ref{sec:learningtodelegate}, here we discuss how to frame subproblem selection as classification, and compare results and discuss trade-offs with regression-based subproblem selection.

\paragraph{Subproblem selection as classification.} The goal of a classification-based subproblem selector is to select the subproblem which results in the best immediate improvement when solved by the subsolver. 

\paragraph{Label transformation.} Optimally, we would like our subproblem selector to select the subproblem with the best immediate improvement. However, since we want to avoid overpenalizing the subproblem selector for putting probability mass on a slightly suboptimal selection, we construct a softmax soft label based on $\delta(S)$ with temperature parameter $\tau$ rather than a one-hot hard label, which corresponds to $\tau \to 0$
\begin{equation}
    \ell(S) = \frac{e^{\delta(S) / \tau}}{\sum_{S' \in \mathcal S_{k,\text{local}}} e^{\delta(S') / \tau}}
\end{equation}

\begin{figure}[ht]
  \centering
  \includegraphics[width=\linewidth]{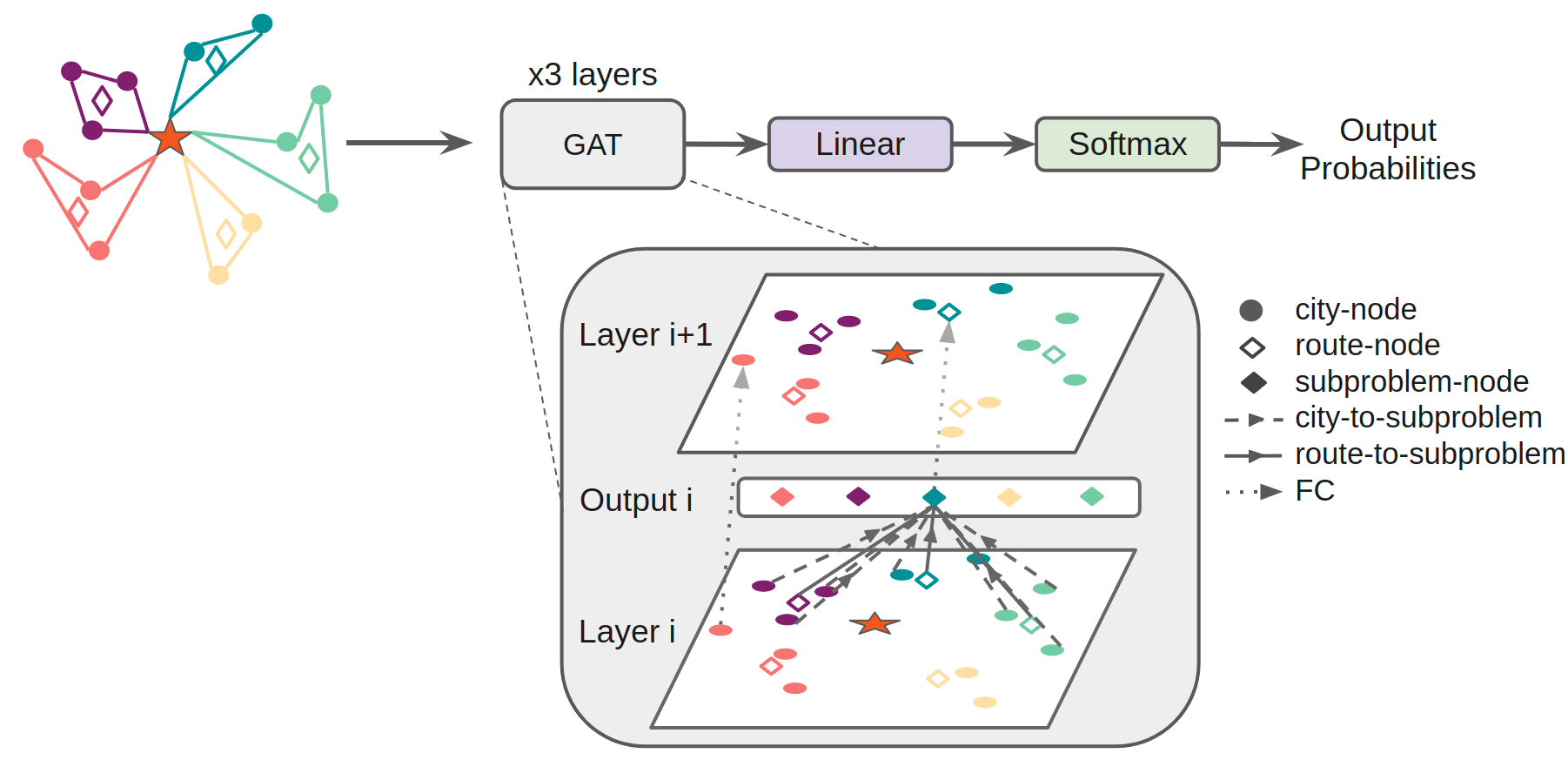}
  \caption{\textbf{Our Graph Attention Network (GAT) architecture}. At each step of our iterative framework (Figure~\ref{fig:pipeline}(c)), we featurize the problem instance $P$ with current solution $X$ into city-node (filled circle) and route-node (empty diamond) features and city-to-subproblem and route-to-subproblem edges. The city-node input features at the $i$th graph attention layer is passed through a FC network to obtain the input city-node features at the $i+1$th graph attention layer. The subproblem-node (filled diamond) output features of the $i$th graph attention layer is passed through a FC network to obtain the input route-node features at the $i+1$th graph attention layer. The final graph attention layer's subproblem-node output features are fed into a linear layers with softmax activation to obtain the classification probabilities $p_\phi(S)$ for each subproblem.}\label{fig:architecture_gat}
\end{figure}

\paragraph{Network architecture.} An important consequence of subproblem classification is that all subproblems must be considered jointly. This requires a different architecture than presented in Section~\ref{sec:learningtodelegate}. Intuitively, in order to consider which subproblem is \textit{best}, the other subproblems must be taken into account. In the regression context, a local measure of improvement is all that is needed. More precisely, as subproblem classification inherently requires feed-forward and back-propagation computation over all subproblems $S \in \mathcal S_{k, \text{local}}$ for a single problem $P$, our choice of architecture must address this constraint to feasibly train with a large enough batch of problems. Unlike in subproblem regression, we cannot simply sample batches of subproblems from the set of all problem instances to fit individually; each computation in classification entails a softmax over $|S_{k, \text{local}}|$ subproblems from the same problem instance. Therefore, our classification-based subproblem selector consists of a Graph Attention Network (GAT) backbone \citep{velickovic2018graph}, which shares computation and memory between overlapping subproblems in the same problem instance, permitting the use of large batches of problem instances $P$ to increase training stability.

We illustrate the nodes and edges of the graph attention in Figure~\ref{fig:architecture_gat}. Given a problem instance $P$, we define three types of graph nodes: one \textit{city-node} per city in $P$, one \textit{route-node} per route $r$ in the current solution $X$, and one subproblem node per $S \in \mathcal S_{k, \text{local}}$. We define two types of directed edge connections: a \textit{city-to-subproblem} edge connects a city-node to a subproblem-node if the city is in the subproblem and a \textit{route-to-subproblem} edge connects a route-node to a subproblem-node if the route is in the subproblem. The input city-node features is fed into a fully connected (FC) network to obtain the input city-node features of the next graph attention layer, while the output subproblem-node features is fed into another FC network to obtain the input route-node features for the next graph attention layer, as each subproblem $S$ corresponds to a single route $r$ by construction as described in Subsection~\ref{sec:subproblem_space}. The subproblem-node features of the final graph attention layer are fed into a linear layer with softmax activation to obtain classification probabilities.

\paragraph{KL divergence loss.} We minimize the KL divergence between the ground-truth soft labels $\ell(S)$ and the subproblem selector output probability distribution $p_\phi(S)$, where $\phi$ is the parameters of the subproblem selector
\begin{equation}
    L(\phi; P) = D_{\text{KL}}(\ell\|p_\phi) = \sum\limits_{S \in \mathcal S_{k, \text{local}}} \ell(S) \log\Big(\frac{\ell(S)}{p_\phi(S)}\Big).
\end{equation}

One possible advantage of the KL divergence loss is that it places the most weight on subproblems with high $\ell(S)$, which may help the neural network focus on a few subproblems with high immediate improvements rather than the many subproblems with low immediate improvement. In contrast, the Huber loss used in subproblem regression places equal emphasis on fitting subproblems with low and high immediate improvements.

\paragraph{Classification setup.} While we previously describe the setup of the regression model in Appendix~\ref{appsec:uniform_setup}, here we describe the additional modifications made in the classification setup compared with the regression setup.
\begin{enumerate}
    \item While we do not need to collect separate training data from the regression setup, we need to process the label to be $\ell(S)$ rather than $\delta(S)$.
    \item In addition to extracting 3-dim feature vector for each city in $P$, we also extract 8-dim feature vector for each route in the current solution $X$ consisting of: the average, max, and min $(x, y)$ locations of cities in the route (shifted by the depot's $(x, y)$ location); the sum of demands of cities in the route; and the route's distance.
    \item We find the best GAT architecture hyperparameters to be a model (hidden) dimension of 128, 1 attention head, and 3 graph attention layers.
    \item As the classification network does not handle multiple problem sizes (e.g. $N = 500$ and $N = 2000$) naturally due to large differences in the number of subproblems, we train each classification model on data from a single problem instance size. However, this is not a fundamental limitation, and future works may explore training a subproblem classification network on multiple problem sizes.
    \item We train our best classification models with batch sizes of $256, 256$, and $512$ \textit{problem instances} respectively for $N = 500, 1000, 2000$. We emphasize that the batches here are batches of \textit{problem instances}, where each problem instance graph inherently contains $|\mathcal S_{k, \text{local}}|$ subproblems as illustrated earlier in Figure~\ref{fig:architecture_gat}, rather than batches of \textit{subproblems} as in the regression setup.
    \item Training takes around 6 hours for $N = 500$, 12 hours for $N = 1000$, and 24 hours for $N = 2000$ on a single NVIDIA V100 GPU.
\end{enumerate}

\FloatBarrier
\begin{table}[ht]
\centering
\caption{\textbf{Training time of subproblem regression and subproblem classification.} Subproblem regression models are simultaneously trained over all problem sizes where subproblem classification models are trained on individual problem sizes. Training times are similar for subproblems with size $k = 5$ and $k = 10$, so we do not distinguish between the two cases in this table. Note that we do not use $N = 2000$ data for subproblems with size $k = 10$ for training, as explained in Appendix~\ref{appsec:uniform_setup}, but we do use $N = 2000$ data for subproblems with size $k = 5$ for training, as explained in Appendix~\ref{appsec:k5k10}.}
\vspace*{0.1cm}
\label{tab:training_cls_reg}
\begin{tabular}{ccc}
\toprule % from booktabs package
\bfseries $N$ & \bfseries Regression & \bfseries Classification\\
\midrule % from booktabs package
$500$ & \multirow{3}{1.5cm}{\centering \textbf{6 hours total}} & 6 hours\\
$1000$ &  & 12 hours\\
$2000$ &  & 24 hours\\
\bottomrule
\end{tabular}
\end{table}
\FloatBarrier

Table~\ref{tab:training_cls_reg} summarizes a key advantage of subproblem regression over subproblem classification: subproblem regression is much faster to train and only requires a single trained model over all problem instance sizes.

\begin{figure}[ht]
  \centering
  \includegraphics[width=\linewidth]{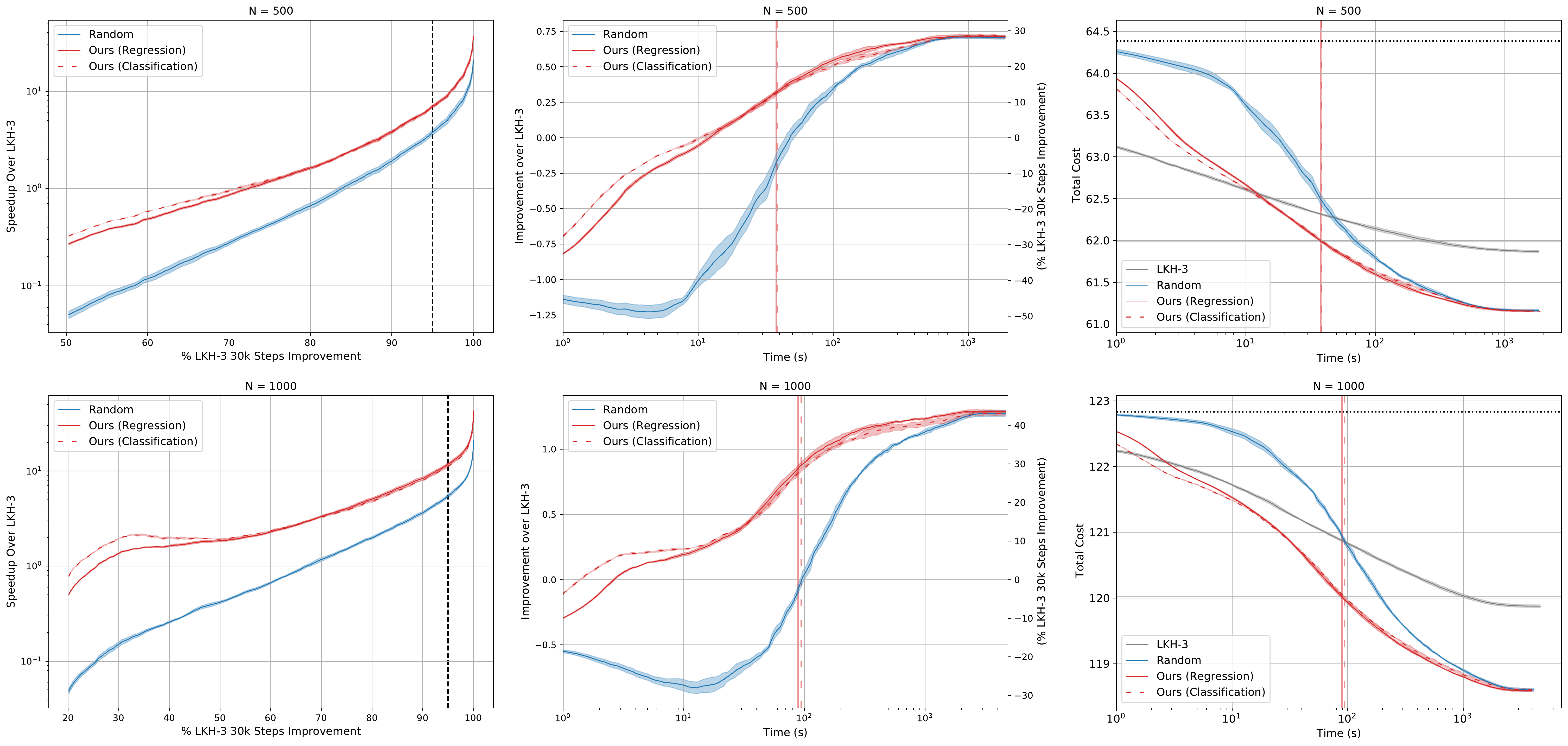}
  \caption{\textbf{Uniform CVRP distribution, regression vs classification.} Graphs of speedup (left column, higher is better), improvement (middle column, higher is better), and total cost (right column, lower is better). The right vertical axis in each improvement graph (middle column) indicates the improvement over LKH-3 as a percentage of the total improvement that LKH-3 attains over 30k steps. We use loosely dashed lines to indicate classification results. The horizontal dotted black line in the total cost graph (right) indicates the initial solution quality for all methods. The vertical dashed black line in the speedup graph (left) and the horizontal LKH-3-colored line in the total cost graph (right) indicate the 95\% LKH-3 solution quality. The styled vertical lines in the improvement graph (middle) and the total cost graph (right) indicate computation times required for the corresponding learning-based methods to reach the aforementioned solution quality.}\label{fig:uniform_reg_cls}
\end{figure}

\paragraph{Comparison on uniform CVRP.} We compare the performance of subproblem regression and subproblem classification for $N = 500$ and $1000$ uniform CVRP and $k = 10$. Despite requiring significantly less training time, we observe that subproblem regression does not suffer a loss in performance against the more specialized subproblem classification, as seen in Figure~\ref{fig:uniform_reg_cls}. Thus, we chose to focus on subproblem regression in this paper as it offers generalization over multiple problem sizes while enjoying significantly less training time.

\begin{figure}[ht]
  \centering
  \includegraphics[width=\linewidth]{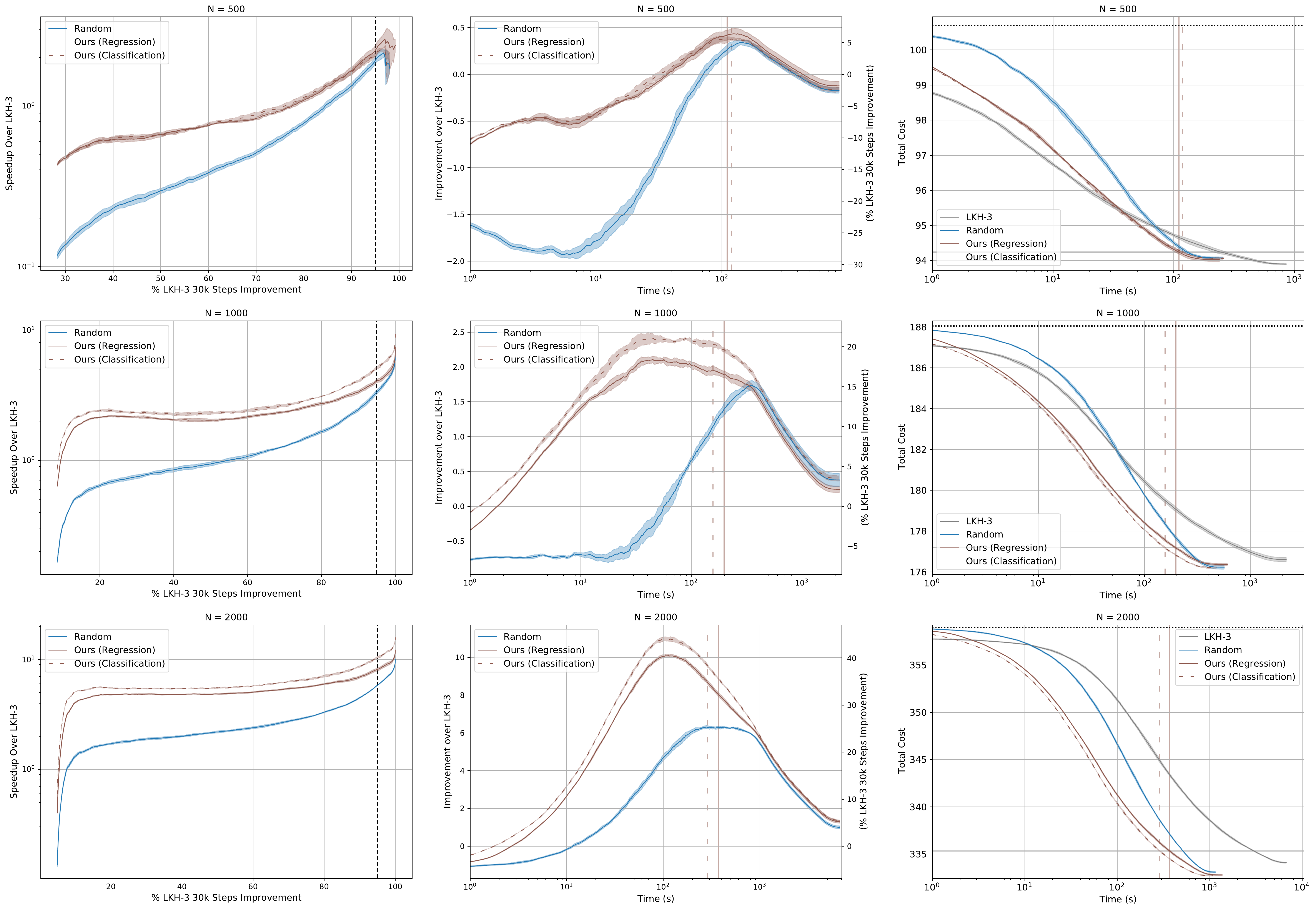}
  \caption{\textbf{Uniform CVRPTW distributions.} Graphs of speedup (left column, higher is better), improvement (middle column, higher is better), and total cost (right column, lower is better). The right vertical axis in each improvement graph (middle column) indicates the improvement over LKH-3 as a percentage of the total improvement that LKH-3 attains over 30k steps. We use loosely dashed lines to indicate classification results. The horizontal dotted black line in the total cost graph (right) indicates the initial solution quality for all methods. The vertical dashed black line in the speedup graph (left) and the horizontal LKH-3-colored line in the total cost graph (right) indicate the 95\% LKH-3 solution quality. The styled vertical lines in the improvement graph (middle) and the total cost graph (right) indicate computation times required for the corresponding learning-based methods to reach the aforementioned solution quality.}\label{fig:cvrptw_reg_cls}
\end{figure}

\begin{figure}[ht]
  \centering
  \includegraphics[width=\linewidth]{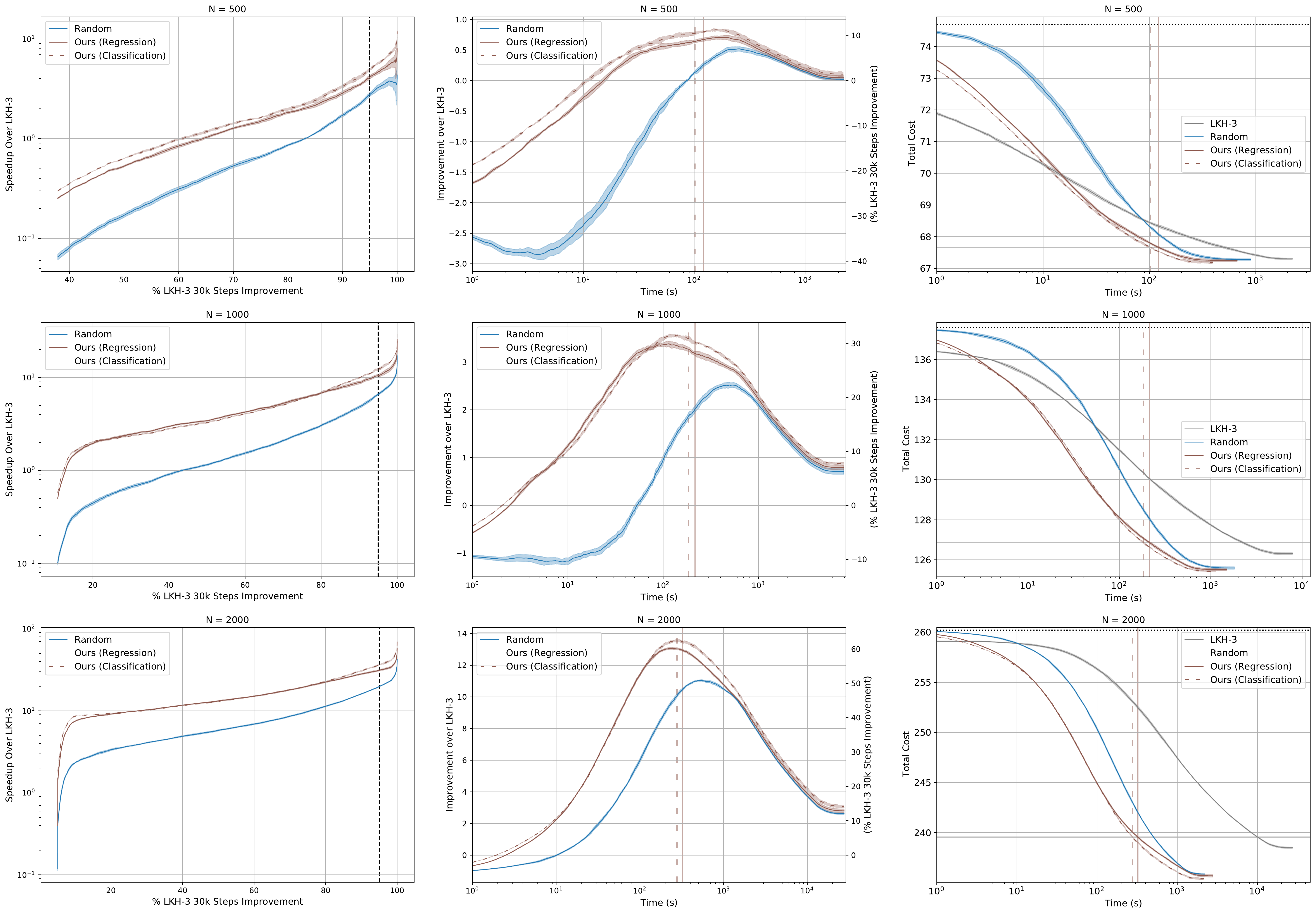}
  \caption{\textbf{Uniform VRPMPD distributions.} Graphs of speedup (left column, higher is better), improvement (middle column, higher is better), and total cost (right column, lower is better). The right vertical axis in each improvement graph (middle column) indicates the improvement over LKH-3 as a percentage of the total improvement that LKH-3 attains over 30k steps. We use loosely dashed lines to indicate classification results. The horizontal dotted black line in the total cost graph (right) indicates the initial solution quality for all methods. The vertical dashed black line in the speedup graph (left) and the horizontal LKH-3-colored line in the total cost graph (right) indicate the 95\% LKH-3 solution quality. The styled vertical lines in the improvement graph (middle) and the total cost graph (right) indicate computation times required for the corresponding learning-based methods to reach the aforementioned solution quality.}\label{fig:vrpmpd_reg_cls}
\end{figure}

\paragraph{Comparison on VRP variants.} We compare the performance of subproblem regression and subproblem classification for CVRPTW and VRPMPD instances of size $N = 500, 1000$ and $2000$ with $k = 5$. Interestingly, we see that subproblem classification somewhat outperforms subproblem regression in Figure~\ref{fig:cvrptw_reg_cls} and Figure~\ref{fig:vrpmpd_reg_cls}. Therefore, future works may benefit from trying both regression and classification and possibly adapting subproblem classification to train over multiple problem sizes.

\subsubsection{Subproblem Size $k$ and Asymptotic Behavior in Uniform CVRP}
\label{appsec:k5k10}
We explore the effect of the subproblem size $k$, which specifies the number of nearest neighbor routes used to create each subproblem, and show that the choice of $k$ may affect the asymptotic behavior of our iterative framework. We compare results between $k = 5$ and $k = 10$, using the same $k = 5$ data generation process as described in Appendix~\ref{app:cvrptw} for uniform CVRP (not CVRPTW).

\begin{figure}[ht]
  \centering
  \includegraphics[width=\linewidth]{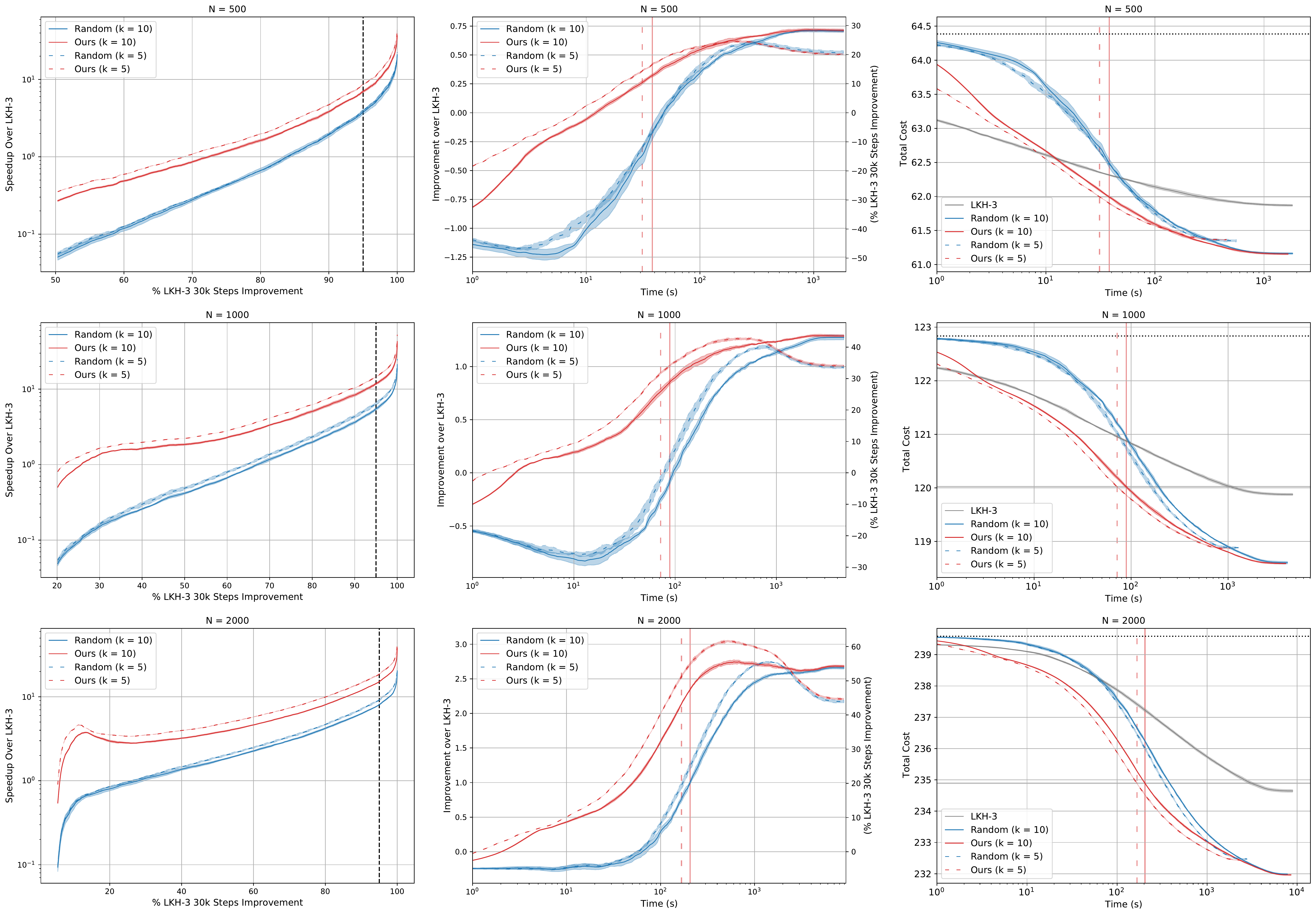}
  \caption{\textbf{Uniform CVRP distributions, $k = 10$ and $k = 5$.} Graphs of speedup (left column, higher is better), improvement (middle column, higher is better), and total cost (right column, lower is better). The right vertical axis in each improvement graph (middle column) indicates the improvement over LKH-3 as a percentage of the total improvement that LKH-3 attains over 30k steps. We use loosely dashed lines to indicate $k = 5$ results. The horizontal dotted black line in the total cost graph (right) indicates the initial solution quality for all methods. The vertical dashed black line in the speedup graph (left) and the horizontal LKH-3-colored line in the total cost graph (right) indicate the 95\% LKH-3 solution quality. The styled vertical lines in the improvement graph (middle) and the total cost graph (right) indicate computation times required for the corresponding learning-based methods to reach the aforementioned solution quality.}\label{fig:k5k10}
\end{figure}

\paragraph{Experimental results.} As shown in Figure~\ref{fig:k5k10}, our method and the Random baseline with $k = 5$ may offer additional speedup over their $k = 10$ counterparts. However, they tend to converge earlier to worse eventual solution qualities when run for a very long time, though still better than the eventual solution qualities of LKH-3. These results suggest that subproblem selection incorporating both $k = 5$ and $k = 10$ could combine the superior speedup of $k = 5$ with the better eventual solution qualities of $k = 10$.

\subsubsection{CVRP Subsolvers: LKH-3 vs HGS}
\label{appsec:subsolver}
We seek to demonstrate the independence of our framework from the LKH-3 subsolver by applying our framework with HGS~\cite{vidal2012hybrid, vidal2020hybrid} as the subsolver to the problem instances from Appendix~\ref{appsec:uniform_setup} and Appendix~\ref{appsec:clustered}. Here we discuss our HGS-related experimental setup and results.

\paragraph{Initialization solution.} To facilitate comparison with previous results, we use the exact same \textit{LKH-3-based} initialization solutions for all instances as described in Appendix~\ref{appsec:uniform_setup}. One note is that the HGS solver does not take an initial solution, so we cannot start running the HGS baseline from the same LKH-3-based initialization as the other methods. To compensate for this limitation, we instead subtract the mean LKH-3-based initialization time from the HGS baseline's total runtime. After this runtime adjustment, we calculate the total HGS improvement by defining the initial HGS solution quality to be the solution quality of our initialization for all other methods. Overall, the initialization time is negligible compared to the total computation time.

\paragraph{Training data generation.} As the HGS subsolver is significantly faster than LKH-3, we run the subsolver for 1 second on each subproblem (compared to 6.7 seconds for LKH-3 on size $k = 10$ subproblems). This allows us to collect data for $N = 500, 1000,$ and $2000$ and $D_\text{train} = 80, 160,$ and $320$, respectively, within 10 hours total on 200 CPUs. The remaining uniform and clustered/mixed setup details are described in Appendix~\ref{appsec:uniform_setup} and Appendix~\ref{appsec:clustered}, respectively. We collect HGS-based training data for both uniform CVRP and clustered/mixed CVRP, then train a subproblem selector for each data distribution.

\paragraph{Baseline.} We compute the speedup and improvement of HGS-based subproblem selectors relative to the HGS solver itself. For uniform CVRP, we run the HGS baseline for the same amount of computation time as our LKH-3 baseline: 1800 seconds, 4620 seconds, 8940 seconds, and 30000 seconds for $N = 500, 1000, 2000,$ and $3000$, respectively. However, since there are multiple combinations of $(N, n_c, m)$ for clustered and mixed distributions with different computation times, we run the HGS baseline for 2000 seconds, 5000 seconds, and 10000 seconds for $N = 500, 1000,$ and $2000$ for every $(N, n_c, m)$.

\begin{figure}[ht]
  \centering
  \includegraphics[width=\linewidth]{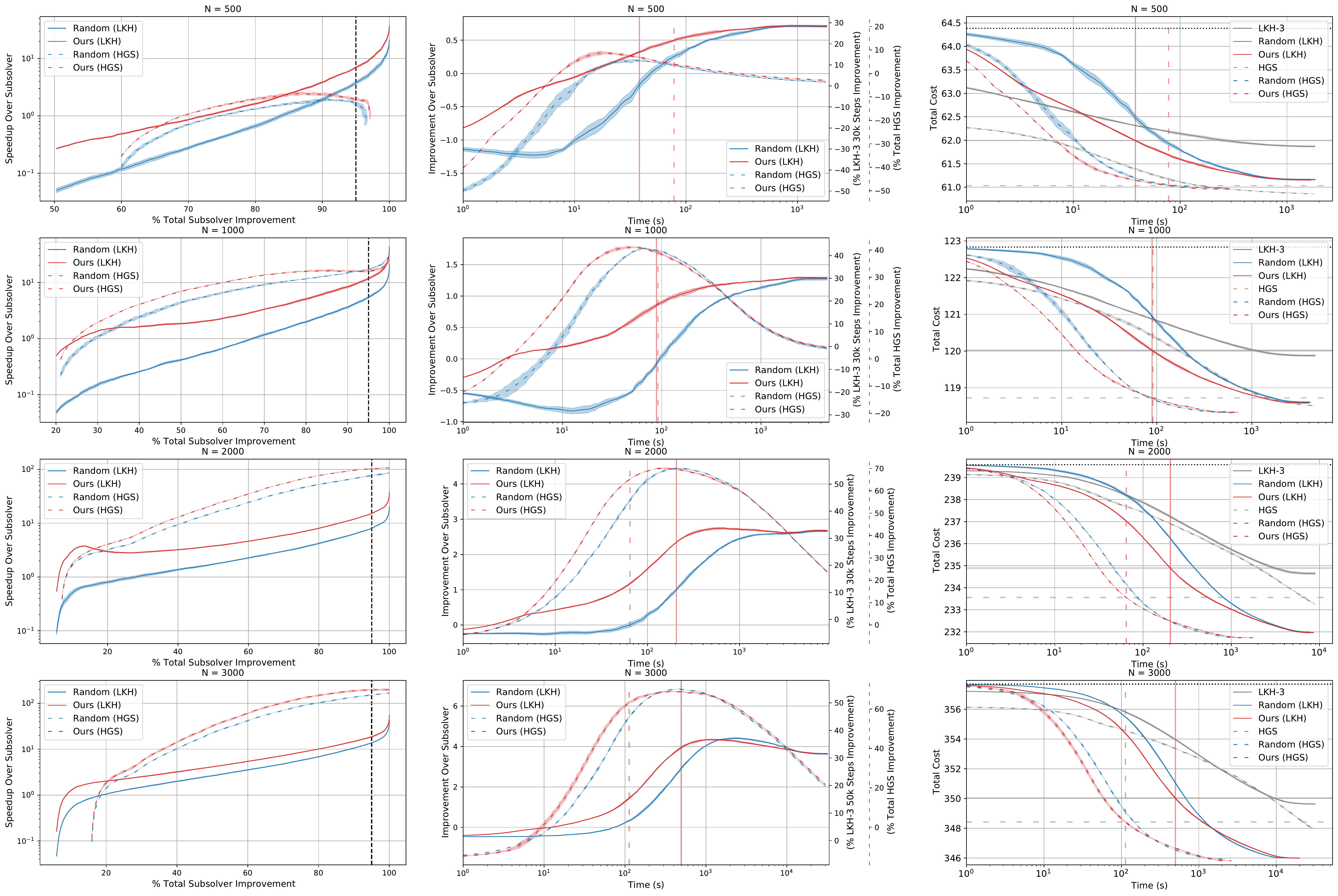}
  \caption{\textbf{Uniform CVRP distributions, LKH-3 and HGS subsolvers.} Graphs of speedup (left column, higher is better), improvement (middle column, higher is better), and total cost (right column, lower is better). We use loosely dashed lines to indicate HGS results. The right vertical axes in each improvement graph (middle column) indicates the improvement over the corresponding subsolver as a percentage of the best total improvement that the subsolver baseline attains. The horizontal dotted black line in the total cost graph (right) indicates the initial solution quality for all methods. The vertical dashed black line in the speedup graph (left) and the horizontal subsolver-colored lines in the total cost graph (right) indicate 95\% solution qualities for the corresponding subsolvers. The styled vertical lines in the improvement graph (middle) and the total cost graph (right) indicate computation times required for the corresponding learning-based methods to reach the corresponding aforementioned solution qualities.}\label{fig:lkh_hgs_uniform}
\end{figure}

\begin{figure}
  \centering
  \includegraphics[width=\linewidth]{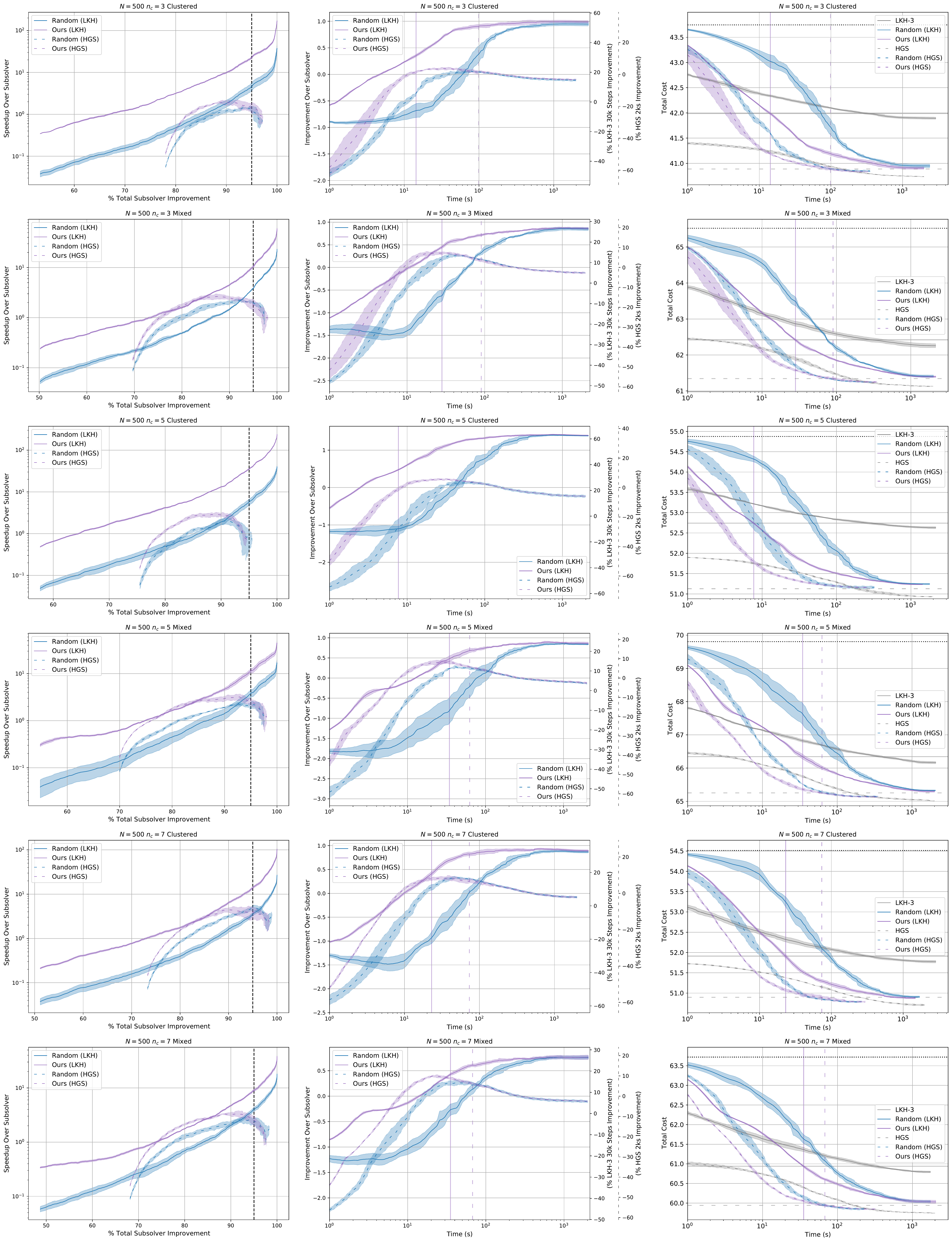}
  \caption{\textbf{Clustered and mixed CVRP distributions, $N = 500$, LKH-3 and HGS subsolvers.} Graphs of speedup (left column, higher is better), improvement (middle column, higher is better), and total cost (right column, lower is better). We use loosely dashed lines to indicate HGS results. The right vertical axes in each improvement graph (middle column) indicates the improvement over the corresponding subsolver as a percentage of the best total improvement that the subsolver baseline attains. The horizontal dotted black line in the total cost graph (right) indicates the initial solution quality for all methods. The vertical dashed black line in the speedup graph (left) and the horizontal subsolver-colored lines in the total cost graph (right) indicate 95\% solution qualities for the corresponding subsolvers. The styled vertical lines in the improvement graph (middle) and the total cost graph (right) indicate computation times required for the corresponding learning-based methods to reach the corresponding aforementioned solution qualities.}\label{fig:cluster_n500}
\end{figure}

\begin{figure}
  \centering
  \includegraphics[width=\linewidth]{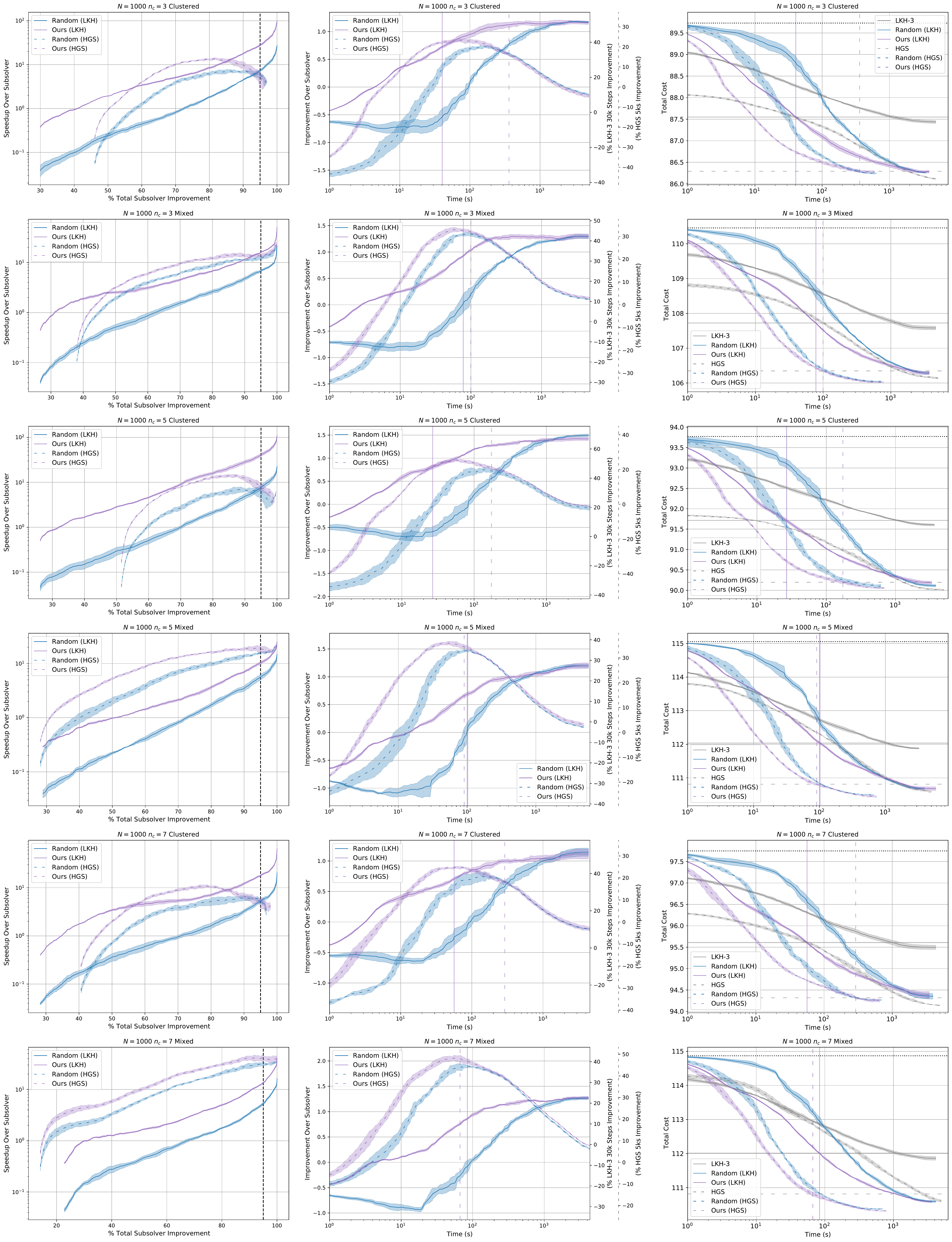}
  \caption{\textbf{Clustered and mixed CVRP distributions, $N = 1000$, LKH-3 and HGS subsolvers.} Graphs of speedup (left column, higher is better), improvement (middle column, higher is better), and total cost (right column, lower is better). We use loosely dashed lines to indicate HGS results. The right vertical axes in each improvement graph (middle column) indicates the improvement over the corresponding subsolver as a percentage of the best total improvement that the subsolver baseline attains. The horizontal dotted black line in the total cost graph (right) indicates the initial solution quality for all methods. The vertical dashed black line in the speedup graph (left) and the horizontal subsolver-colored lines in the total cost graph (right) indicate 95\% solution qualities for the corresponding subsolvers. The styled vertical lines in the improvement graph (middle) and the total cost graph (right) indicate computation times required for the corresponding learning-based methods to reach the corresponding aforementioned solution qualities.}\label{fig:cluster_n1000}
\end{figure}

\begin{figure}
  \centering
  \includegraphics[width=\linewidth]{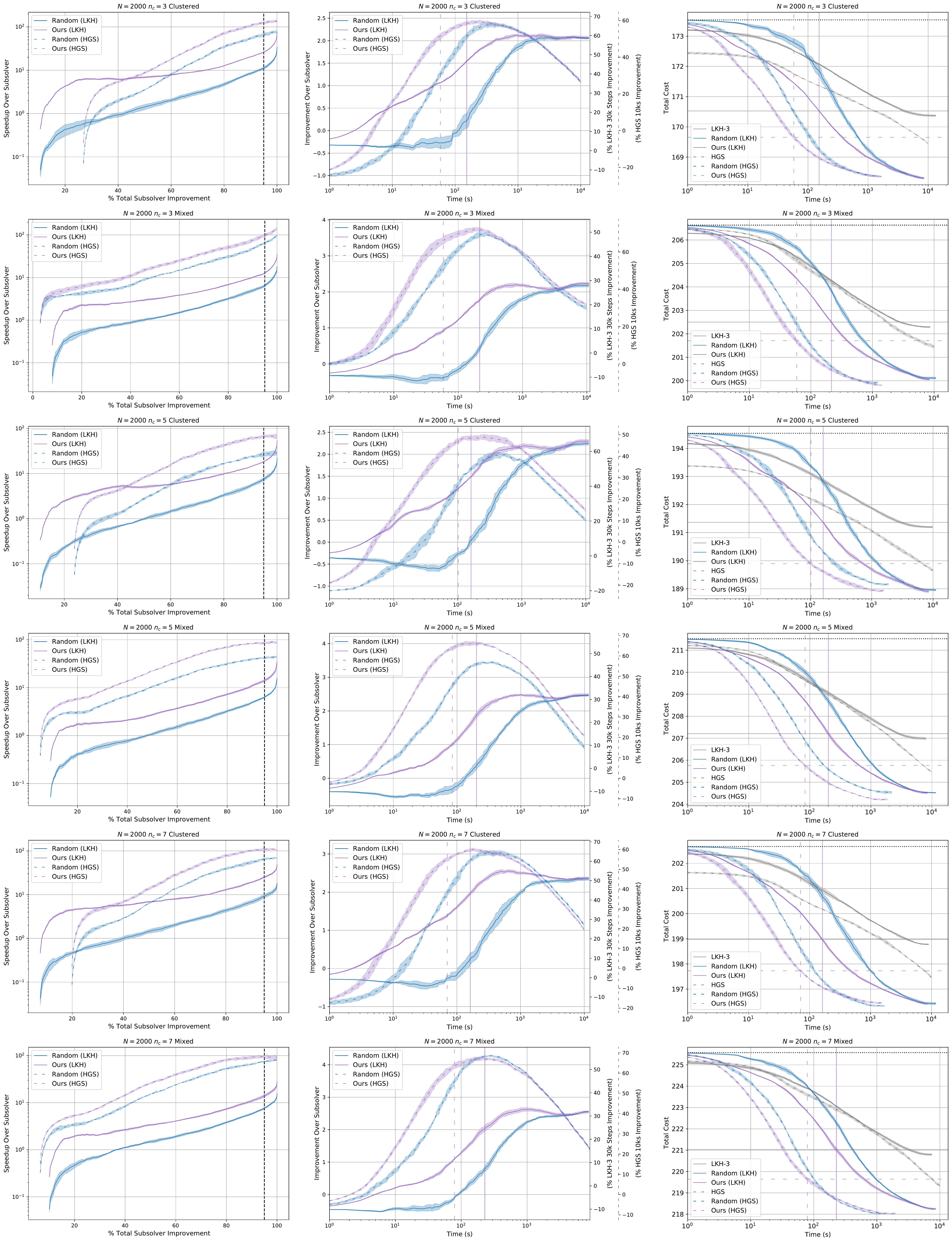}
  \caption{\textbf{Clustered and mixed CVRP distributions, $N = 2000$, LKH-3 and HGS subsolvers.} Graphs of speedup (left column, higher is better), improvement (middle column, higher is better), and total cost (right column, lower is better). We use loosely dashed lines to indicate HGS results. The right vertical axes in each improvement graph (middle column) indicates the improvement over the corresponding subsolver as a percentage of the best total improvement that the subsolver baseline attains. The horizontal dotted black line in the total cost graph (right) indicates the initial solution quality for all methods. The vertical dashed black line in the speedup graph (left) and the horizontal subsolver-colored lines in the total cost graph (right) indicate 95\% solution qualities for the corresponding subsolvers. The styled vertical lines in the improvement graph (middle) and the total cost graph (right) indicate computation times required for the corresponding learning-based methods to reach the corresponding aforementioned solution qualities.}\label{fig:cluster_n2000}
\end{figure}

\paragraph{Experimental results.} Figure~\ref{fig:lkh_hgs_uniform} compares the speedup over the subsolver when using either LKH-3 or HGS as the subsolver for uniform CVRP of size $N = 500, 1000, 2000,$ and $3000$. Similarly, Figures~\ref{fig:cluster_n500},~\ref{fig:cluster_n1000},~and~\ref{fig:cluster_n2000} contain results for mixed and clustered CVRP distributions of size $N = 500, 1000,$ and $2000$, respectively. We observe that for $N = 500$, our method with the HGS subsolver offers a small speedup over the HGS solver for attaining intermediate range of solution qualities; however, the HGS solver converged to better final solution quality in all cases. This observation is unsurprising, as HGS is designed and calibrated to obtain state-of-the-art solutions for instances of size 500 to 1000. Nevertheless, as the problem size $N$ increases, the speedup of our method over HGS increases, demonstrating the effectiveness of subproblem selection in large-scale problems. For $N = 2000$ and $3000$ we are not able to run HGS until full convergence, which would require another order of magnitude more time. Our learned method outperformed the Random baseline in all cases, demonstrating that learning further accelerates subproblem selection.


\subsubsection{Effect of Initial Solution Quality}
\label{appsec:initialization}
As our iterative framework relies on an initial solution generated according to Appendix~\ref{appsec:uniform_setup}, we explore the effect of poorer, out-of-distribution initial solutions on our subproblem selector for uniform CVRP instances.

\paragraph{Initialization solution.} Let $L$ denote the number of LKH steps run on each partition at initialization. Previously, $L = 100$ as stated in Appendix~\ref{appsec:uniform_setup}. To generate initial solutions of poorer quality, we additionally generate initial solutions with $L = 1, 5, 10,$ and $50$. As a last initialization method, which does not use LKH-3 and is denoted $L = 0$, we uniformly randomly chain cities into routes with no consideration for their location. We do not train or finetune any models for this ablation, instead evaluating the performance of our previous uniform CVRP model (trained on $L = 100$ initializations) on all other initialization methods. Note that we calculate the speedup and improvement of all methods with respect to LKH-3 with $L = 100$ initialization.

\begin{figure}[ht]
  \centering
  \includegraphics[width=\linewidth]{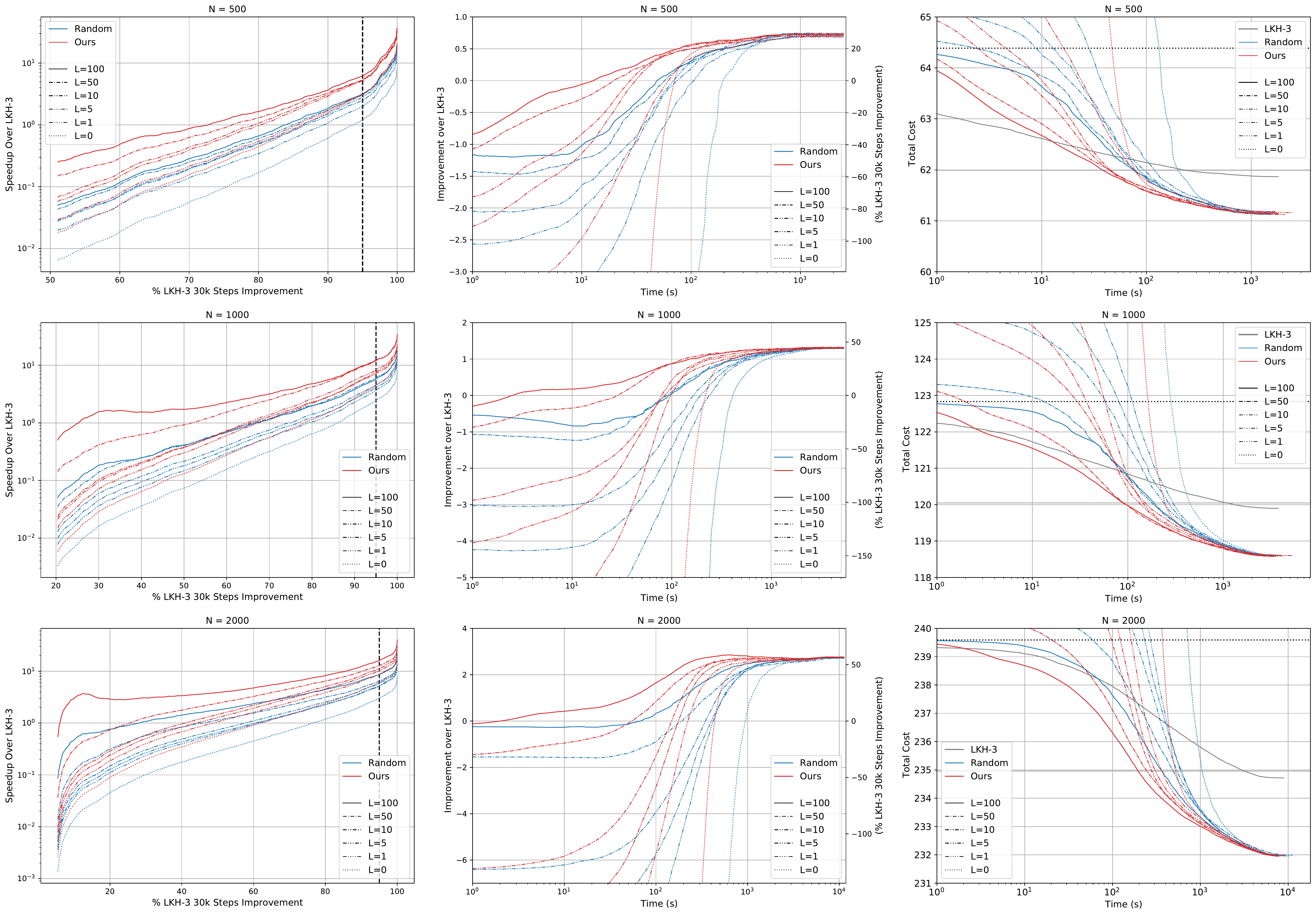}
  \caption{\textbf{Uniform CVRP distributions, various initialization qualities.} Graphs of speedup (left column, higher is better), improvement (middle column, higher is better), and total cost (right column, lower is better). The right vertical axis in each improvement graph (middle column) indicates the improvement over LKH-3 as a percentage of the total improvement that LKH-3 attains over 30k steps. We use increasingly dotted lines to indicate increasingly worse initialization methods. The horizontal dotted black line in the total cost graph (right) indicates the initial solution quality for all methods. The vertical dashed black line in the speedup graph (left) and the horizontal LKH-3-colored line in the total cost graph (right) indicate solution qualities equal to 95\% of the corresponding best LKH-3 solution quality. The styled vertical lines in the improvement graph (middle) and the total cost graph (right) indicate computation times required for the corresponding learning-based methods to reach the corresponding aforementioned solution quality.}\label{fig:initialization_ablation}
\end{figure}

\paragraph{Experimental results.} We provide experimental results in Figure~\ref{fig:initialization_ablation}. We find that for each subproblem selection method (Ours and Random), worse initialization corresponds to somewhat worse speedup and worse improvement over LKH-3 with $L = 100$. Nevertheless, Ours outperforms Random for every $L$ considered, demonstrating that our subproblem selector trained on $L = 100$ data generalizes well to out-of-distribution initial solutions. We are also able to show that subproblem selection with even the worst initialization offers a speedup above 1x for attaining 95\% of $L = 100$ LKH-3 solution quality.

\subsubsection{Transformer vs Simpler Architectures}
\label{appsec:architecture_ablation}
We seek to illustrate the importance of our Transformer regression architecture illustrated in Figure~\ref{fig:architecture} by comparing with subproblem selectors with simpler architectures trained on uniform CVRP.

\paragraph{Experimental setup.} We describe several simpler architectures, some of which require changes to the input subproblem representation.\begin{itemize}
    \item \textbf{FCNN}: using the same data (positions and demands of cities) as described in Section~\ref{appsec:uniform_setup}, we replace the Transformer attention layers by a fully-connected neural network.
    \item \textbf{Linear, MLP, RandomForest}: we design a new input representation which featurizes each subproblem into 33 summary features, including the number of cities, the bounds of the subproblem, the spread of cities, 10 radial distance percentiles of cities from the depot, 10 radial distance percentiles of cities from the subproblem centroid, and the distribution of city demands. On top of this input representation, we fit simple regression models: linear regression (Linear), a shallow fully-connected neural network (MLP), and a Random Forest regressor (RandomForest).
\end{itemize}
To train the simpler subproblem selectors, we use the same instances and training data as training the Transformer subproblem selector. Hyperparameters were lightly tuned by hand.

\begin{figure}[ht]
  \centering
  \includegraphics[width=\linewidth]{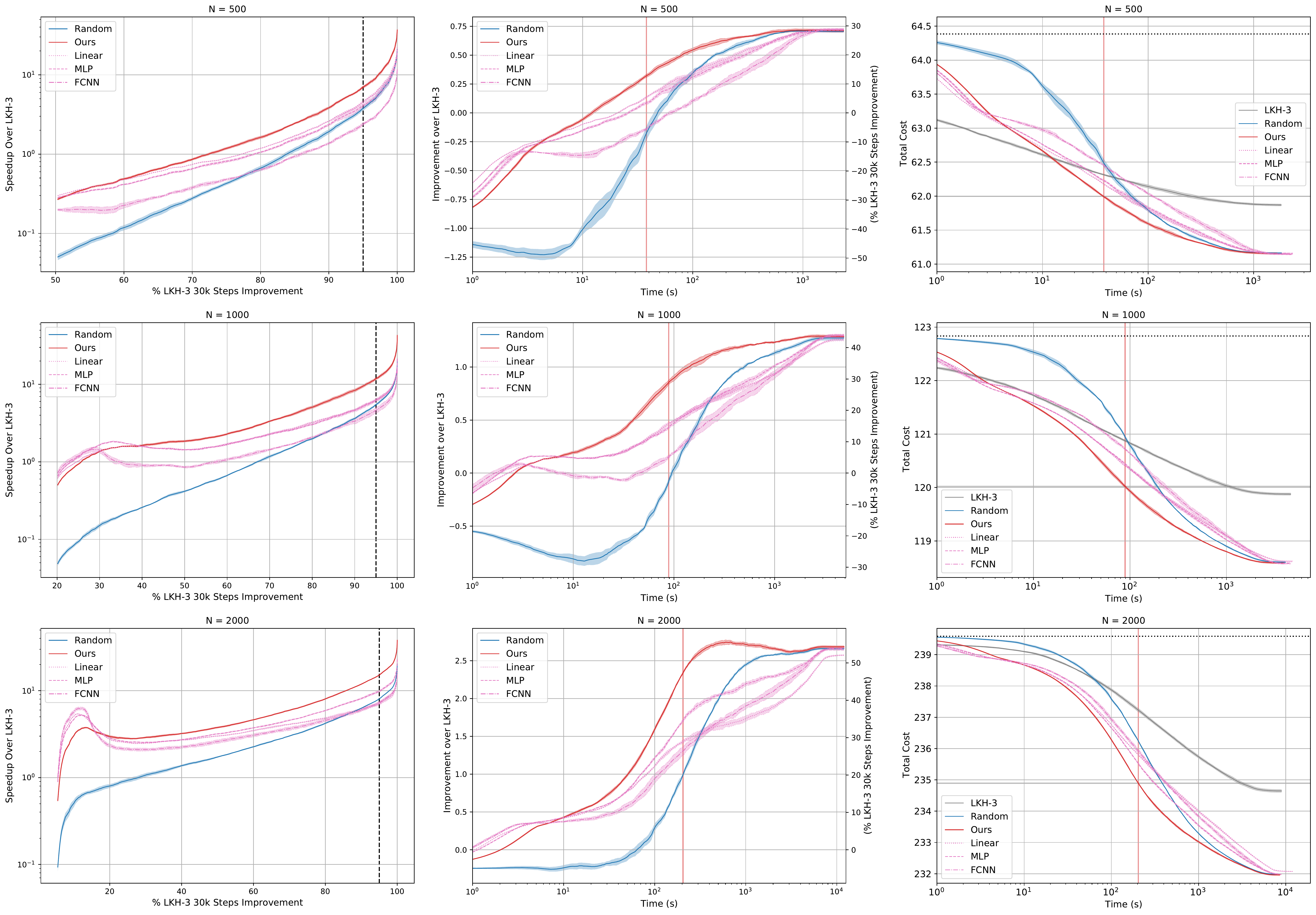}
  \caption{\textbf{Uniform CVRP distributions, ablation architectures.} Graphs of speedup (left column, higher is better), improvement (middle column, higher is better), and total cost (right column, lower is better). The right vertical axis in each improvement graph (middle column) indicates the improvement over LKH-3 as a percentage of the total improvement that LKH-3 attains over 30k steps. We indicate ablation architectures in pink. The horizontal dotted black line in the total cost graph (right) indicates the initial solution quality for all methods. The vertical dashed black line in the speedup graph (left) and the horizontal LKH-3-colored line in the total cost graph (right) indicate solution qualities equal to 95\% of the corresponding best LKH-3 solution quality. The styled vertical lines in the improvement graph (middle) and the total cost graph (right) indicate computation times required for the corresponding learning-based methods to reach the corresponding aforementioned solution quality.}\label{fig:architecture_ablation}
\end{figure}

\paragraph{Experimental results.} The validation mean squared error for ablation architectures are similar to each other (ranging from 0.02 to 0.03) and much higher than that of our Transformer architecture (0.003). Due to the large CPU memory consumption of RandomForest at inference time and its similar validation mean squared error to the other ablation architectures, we do not evaluate RandomForest's subproblem selection ability. Figure~\ref{fig:architecture_ablation} demonstrate the performance of our iterative framework with other subproblems selectors. Interestingly, we see that ablation architectures initially, for roughly the first 5 seconds, perform comparably to our Transformer architecture. However, the quality of the solution gradually diverges from our Transformer architecture and converges to the Random baseline over the first 100 to 400 seconds. MLP, which uses summary feature inputs, performs the best out of all three ablations, though Linear offers similar performance on $N = 500$ and $1000$. Our studies show that our Transformer architecture is key for obtaining good performance over the Random baseline, especially for obtaining higher solution qualities, though summary features are surprisingly effective for subproblem selection even when combined with the interpretable Linear architecture.

\newpage
\subsection{Applicability of Our \textit{learning-to-delegate} Framework to Other Problems in Combinatorial Optimization (CO)}\label{appsec:applicability}
First, our method can be seen as a learning approach to accelerate the broadly applicable POPMUSIC framework on large-scale CO problems (discussed in the related work,  Section~\ref{sec:related_work}), where we learn principled improvement-based criteria for subproblem selection, rather than relying on random or heuristic subproblem selection. In short, as long as a restricted subproblem space can be defined, we expect an existing subsolver can be accelerated with our learning-to-delegate method. We thus give a few such example CO problems:

\begin{itemize}
    \item \textbf{Traveling salesman problem (TSP):} in the closely related TSP problem, a restricted subproblem selection space can be defined as subpaths of a fixed length in the current solution path, as done in~\cite{taillard2019popmusic, helsgaun2018using}. Then, at each iteration, the solution for the unselected path segment is held fixed, while the selected subpath is finetuned by a subsolver.
    \item \textbf{Other graph problems:} similarly, the general POPMUSIC framework is also applied to graph problems such as map labeling (max independent set)~\cite{lalla2016popmusic}, where each subproblem can be defined as the k-hop neighborhoods of a centroid node, which is also a linear subproblem selection space. Naturally, the POPMUSIC framework should also apply to related graph problems such as graph cut and minimum vertex cover.
    \item \textbf{Other problems with spatial or temporal locality:} similar to our subproblem space definition, problems with spatial or temporal locality can leverage this property to reduce the space. Such examples include berth allocation (which closely relates to job scheduling)~\cite{taillard2003heuristic} and p-median clustering~\cite{santini2021decomposition}.
\end{itemize}
The concrete examples given here provide a wide array of important and large-scale CO problems which we expect to benefit from learning-to-delegate. Although our method evaluation focuses on VRP, the key contribution of our method is that, given a subproblem space of tractable size, it can learn a selection strategy to drastically accelerate a subsolver, by pinpointing promising subproblems to increase the objective. On the other hand, the previous methods~\cite{taillard2019popmusic, helsgaun2018using, lalla2016popmusic, taillard2003heuristic, santini2021decomposition} rely on random subproblem selection or simple heuristics to select the subproblems, which expends a large amount of computation solving subproblems that already have good subsolutions.

Second, our method also has the potential to apply to CO problems with a larger subproblem space. For instance, augmenting our method with sampling-based strategies could be promising. We could consider training our subproblem regression network by \textbf{sampling} a number of subproblems at each step to generate a training set. To generate full training trajectories across multiple steps, we can 1) naively choose the greedy optimal subproblem, or 2) use an \textbf{active learning} approach to balance exploration (understanding what kinds of subproblems are promising) with exploitation (improving the objective). While 1) is a straightforward application of our current framework, 2) is an interesting extension that is worthy of further exploration.

Finally, we note that although we focus on VRPs, we design the evaluation of the approach with an eye towards generality. Specifically, VRPs have a rich family of variations, including CVRP, CVRPTW, VRPMPD, as well as different data distributions (among many others), which allows us to demonstrate a degree of generality of the method. Although they are related, each of these problem modifications considerably changes the nature of the underlying CO problem; for example, CVRPTW exhibits an element of scheduling, whereas (C)VRP does not.

\end{document}